\documentclass[lettersize,journal]{IEEEtran} 
\pdfoutput=1
\usepackage{amsmath,amsfonts}
\usepackage{algorithmic}
\usepackage{algorithm}
\usepackage{array}
\usepackage[caption=false,font=normalsize,labelfont=sf,textfont=sf]{subfig}
\usepackage{textcomp}
\usepackage{stfloats}
\usepackage{url}
\usepackage{verbatim}
\usepackage{graphicx}
\usepackage{cite}
\usepackage{booktabs}
\usepackage[super]{nth}
\usepackage{subcaption}
\usepackage{makecell}
\usepackage[left]{lineno} 
\usepackage{xcolor}

\usepackage{hyperref}
\usepackage{xr-hyper}
\begin{document}


\title{Process-Guided Concept Bottleneck Model}

\author{Reza M. Asiyabi, Sam Harrison, John L. Godlee, David Milodowski, Nicole H. Augustin, Penelope J. Mograbi, Timothy R. Baker, Lorena M. Benitez, Samuel J. Bowers, Thomas K. Brade, Joao M. B. Carreiras, Duncan M. Chalo, Vera De Cauwer, Kyle G. Dexter, Hermane Diesse, Mathias I. Disney, Luisa F. Escobar-Alvarado, Manfred Finckh, Tatenda Gotore, Gabriele C. Hegerl, John N. Kigomo, Fainess C. Lumbwe, Francisco Maiato, Rudzani A. Makhado, Collins W. Masinde, Musingo Tito E. Mbuvi, Iain M. McNicol, Edward T.A. Mitchard, Buster P. Mogonong, Wilson A. Mugasha, Aristides Baptista Muhate, Hinji Mutondo, Leena Naftal, Paula Nieto-Quintano, Elifuraha Elisha Njoghomi, Catherine L. Parr, Oliver L. Phillips, Pierre Proces, Tshililo Ramaswiela, Jayashree Ratnam, Mathew Rees, Rasmus Revermann, Natasha Ribeiro, Mahesh Sankaran, Abel M. Siampale, Stephen Sitch, Kathleen G. Smart, Hemant G. Tripathi, Wayne Twine, Gabriel I.K. Uusiku, Helga van der Merwe, Chemuku Wekesa, Benjamin J. Wigley, Mathew Williams, Ellie Wood, Emily Woollen, Shaun Quegan, Steven Hancock, and Casey M. Ryan. 

\thanks{This study is supported by the UK NCEO (National Centre for Earth Observation; UKRI NERC grant NE/R016518/1), and SECO (Resolving the current and future carbon dynamics of the dry tropics; UKRI NERC grants NE/T01279X/1, NE/T012722/1). Plot data are provided by the SEOSAW Partnership (https://seosaw.github.io, NERC grant NE/P008755). H. Diesse was supported by the Fostering Research \& Intra-African Knowledge Transfer Through Mobility \& Education. M. Finckh was supported by the German Federal Ministry of Education and Research in the framework of the TFO (grant no. 01LL0912A) and SASSCAL (grant no.01LG1201J) projects. T. Gotore was supported by Oppenheimer Generations Research and Conservation under the Future Ecosystems for Africa project. E. Mitchard and P. Nieto-Quintano's fieldwork were supported by the US Forest Service and the Wildlife Conservation Society (WCS). B. Mogonong is supported by the SAEON-EFTEON Research Infrastructure and the SAEON Arid Lands Node. M. Rees was supported by NERC through an E4 DTP studentship (NE/S007407/1). M. Sankaran was supported by the National Centre for Biological Sciences, TIFR, India (DAE, GoI grant no. 12-R\&D-TFR-5.04-0800), National Geographic Society (grant 982815) and NERC, UK (grant NE-E017436-1). E. Wood was supported by a National Geographic Early Career Grant (EC-61519R-19) and the Elizabeth Sinclair Irvine Bequest. E. Woollen acknowledges the ACES project (NE/K010395/1) for funding the Mozambique plots. We thank Sally Archibald, Sara Banda, Emanuel Chidumayo, Antonio Valter Chisingui, Jason Donaldson, Barend Erasmus, Rhett Harrison, Miya Kabajani, Vivian Kathambi, Anderson Muchawona, Jonathan Muledi, Toby Pennington, Rose Pritchard, Mylor Ngoy Shutcha, Ifo Averti Suspense, and Jose João Tchamba for their contribution to plot data collection. The authors acknowledge the use of AI-assisted tools for language refinement and editorial feedback during the preparation of this manuscript. All scientific content, methodological development, experiments, and interpretations were conducted by the authors.}

\thanks{Corresponding author: Reza M. Asiyabi (reza.asiyabi@ed.ac.uk)}

\thanks{R. Asiyabi, S. Harrison, J. Godlee, D. Milodowski, P. Mograbi, L. Benitez, S. Bowers,  T. Brade, J. Carreiras, L. Escobar-Alvarado, G. Hegerl, I. McNicol, E. Mitchard, P. Nieto-Quintano, M. Rees, M. Williams, E. Wood, E. Woollen, S. Hancock, and C. Ryan are with the School of GeoSciences, University of Edinburgh, UK.

R. Asiyabi, M. Disney, M. Williams, S. Quegan, and S. Hancock are with the UK National Centre for Earth Observation (NCEO).

N. Augustin is with the School of Mathematics, University of Edinburgh, UK.

P. Mograbi, T. Gotore, C. Parr, and W. Twine are with the School of Animal, Plant and Environmental Sciences, University of the Witwatersrand, South Africa.

T. Baker and O. Phillips are with the School of Geography, University of Leeds, UK.

D. Chalo is with the Dept. of Biology, University of Nairobi, Kenya.

V. Cauwer, H. Diesse, L. Naftal, and G. Uusiku are with Namibia University of Science and Technology, Namibia.

K. Dexter is with the Dept. of Life Sciences and Systems Biology, University of Turin, Italy.

K. Dexter and M. Rees are with the Royal Botanic Garden Edinburgh, UK.

M. Disney is with the Dept. of Geography, University College London, UK.

M. Finckh and R. Revermann are with the Institute of Plant Science and Microbiology, University of Hamburg, Germany.

 J. Kigomo, C. Masinde, M. Mbuvi, and C. Wekesa are with Kenya Forestry Research Institute, Kenya.

F. Lumbwe, and H. Mutondo are with WeForest, Zambia.

F. Maiato is with Universidade Mandume Ya Ndemufayo, \& Herbário do Lubango, ISCED-Huíla, Angola.

R. Makhado is with the Dept. of Biodiversity, University of Limpopo, South Africa.

B. Mogonong, T. Ramaswiela, K. Smart, and H. van der Merwe are with the South African Environmental Observation Network (SAEON), South Africa.

W. Mugasha is with the Dept. of Forest Resources Assessment and Management, Sokoine University of Agriculture, Tanzania.

A. Muhate is with the Ministry of Agriculture, Environment and Fisheries, Mozambique.

E. Njoghomi is with the Forestry Research Institute, Tanzania.

C. Parr is with the School of Environmental Sciences, University of Liverpool, UK \& Dept. of Zoology and Entomology, University of Pretoria, South Africa.

P. Proces is with Nature+, Belgium.

J. Ratnam and M. Sankaran are with the National Centre for Biological Sciences (TIFR), India.

N. Ribeiro is with the Dept. of Forest Engineering, Eduardo Mondlane University, Mozambique.

A. Siampale is with the World Wide Fund for Nature (WWF), Zambia.

S. Sitch is with the Faculty of Environment, Science and Economy, University of Exeter, UK.

H. Tripathi is with the Faculty of Biological Sciences, University of Leeds, UK.

H. van der Merwe is with the Dept. of Biological Sciences, University of Cape Town, South Africa.

B. Wigley is with SANParks, South Africa \& School of Natural Resource Management, Nelson Mandela University, South Africa \& University of Bayreuth, Germany.

S. Quegan is with the School of Mathematical and Physical Sciences, University of Sheffield, UK.
}}


\markboth{IEEE TRANSACTIONS ON PATTERN ANALYSIS AND MACHINE INTELLIGENCE, Vol.~V, No.~N, January~2026}%
{Shell \MakeLowercase{\textit{et al.}}: A Sample Article Using IEEEtran.cls for IEEE Journals}


\maketitle

\begin{abstract}
Concept Bottleneck Models (CBMs) improve the explainability of black-box Deep Learning (DL) by introducing intermediate semantic concepts. However, standard CBMs often overlook domain-specific relationships and causal mechanisms, and their dependence on complete concept labels limits applicability in scientific domains where supervision is sparse but processes are well defined. To address this, we propose the Process-Guided Concept Bottleneck Model (PG-CBM), an extension of CBMs which constrains learning to follow domain-defined causal mechanisms through biophysically meaningful intermediate concepts. Using above ground biomass density estimation from Earth Observation data as a case study, we show that PG-CBM reduces error and bias compared to multiple benchmarks, whilst leveraging multi-source heterogeneous training data and producing interpretable intermediate outputs. Beyond improved accuracy, PG-CBM enhances transparency, enables detection of spurious learning, and provides scientific insights, representing a step toward more trustworthy AI systems in scientific applications.
\end{abstract}

\begin{IEEEkeywords}
Above Ground Biomass Density, Artificial Intelligence, Deep Learning, Earth Observation, Explainable AI, Physics-Informed Deep Learning, Process-Guided AI.
\end{IEEEkeywords}

\section{Introduction}
\IEEEPARstart{D}{eep} Learning (DL) has achieved remarkable success across many scientific domains, \cite{tuia2024artificial, xiong2024earthnets}. Yet, its adoption in high-stakes applications remains limited due to critical shortcomings \cite{datcu2023explainable}. Standard DL models often behave as black-box systems, optimised purely for accuracy without considering underlying biophysical laws. As a result, they lack interpretability and consistency with scientific processes \cite{datcu2023explainable, reichstein2019deep, raissi2019physics}, struggle to learn from sparse and heterogeneous annotated training data \cite{hua2021semantic, xiao2025foundation}, and are prone to spurious correlations rather than capturing true mechanisms \cite{geirhos2020shortcut, chai2025identifying}. These limitations reduce trust and hinder the use of AI in sensitive domains such as ecological monitoring, carbon accounting, and related areas of policy-making.

A promising approach to address the interpretability issue is Concept Bottleneck Models (CBMs) \cite{pmlr-v119-koh20a}, where latent bottleneck representations are mapped to human-understandable concepts. A user can inspect the bottleneck concepts to interpret the decision pathway of the DL model and if they disagree, they can intervene and adjust the concept value, which in turn affects the final output \cite{pmlr-v119-koh20a}.

However, vanilla CBMs, as formulated in \cite{pmlr-v119-koh20a}, require fine-grained input ($x_i$), concept ($z_i$), and output ($y_i$) annotation, which limits their applicability in use cases with sparsely annotated data, where each input would have the annotation label for only some of the intermediate concepts and only some of the samples would have the final output label. Moreover, in many scientific applications, there is a well understood process (e.g., biophysical systems) that links the input data to the output through inter-correlated intermediate attributes that often reflect the direction of causality. These intermediate attributes can be seen as the bottleneck concepts, but vanilla CBMs do not consider the underlying biophysical correlation between them and the scientific process that connect the input to the final output via the intermediate attributes. 

In this study, we propose Process-Guided Concept Bottleneck Models (PG-CBMs), an extension of CBMs that guides the DL model through a domain-specific process and embeds the domain knowledge directly into the architecture by constraining intermediate concepts to reflect meaningful causal processes (e.g., ecological mechanisms). In this way, the model is guided along biophysically plausible pathways, enhancing interpretability, reducing the risk of spurious learning, and enabling training with heterogeneous supervision sources. The main contributions of this work are:

\begin{itemize}
    \item Introducing PG-CBM, an extension of CBM that embeds domain-specific causal processes into the bottleneck, ensuring that intermediate representations correspond to mechanistic variables rather than arbitrary semantic tags.
    \item Extending CBMs to handle heterogeneous and partially overlapping supervision sources, enabling effective learning in sparse and noisy scientific datasets where complete $(x,z,y)$ tuples are unavailable.
    \item Accounting for causal interdependencies among intermediate variables by jointly optimising their mappings and constraining the aggregation function $g(\cdot)$ to respect known process-based relationships.
    \item Providing a theoretical justification for improved Out-of-Distribution (OOD) generalisation, showing that process-guided constraints act as a structural regulariser on the hypothesis space.
    \item Demonstrating empirically that PG-CBM retains the predictive accuracy of black-box DL models while offering interpretable, causally consistent intermediate outputs that support model diagnosis and scientific insight.
\end{itemize}

To evaluate the proposed PG-CBM in a scientific domain with a well-defined process-based structure, we apply it to the task of Above Ground Biomass Density (AGBD) estimation from Earth Observation (EO) data. AGBD is a fundamental ecological variable that describes the structure of ecosystems and their role in mass and energy fluxes. AGBD is an Essential Climate Variable of the Global Climate Observing System (GCOS) and is also critical for carbon accounting, but it is difficult to measure directly and therefore requires models that are trustworthy, interpretable, and robust under sparse supervision. AGBD estimation is a particularly well-suited use case for evaluating PG-CBM because:
\begin{itemize}
    \item In ecological practice, above ground biomass is unfeasible to directly measure, even in situ, and so is estimated indirectly using allometric scaling relationships (well understood empirical processes) between measurable attributes such as height and stem diameter, and AGBD.
    \item EO data, particularly estimates of radar backscatter from Synthetic Aperture Radar (SAR) sensors, do not measure biomass directly, but they do contain information about structural attributes of the vegetation that correlate with AGBD in context-specific ways \cite{woodhouse2012radar}. A well-defined model should consider the causal mechanisms that lead to the observed radar backscatter and the associated ecological relationships if it is to successfully infer AGBD from EO data.
    \item Labels for these intermediate attributes and the final AGBD values are not consistently available for all samples, since they originate from different sources (see~\ref{section_dataset} for details).
\end{itemize}

Our experiments show that PG-CBM outperforms black-box DL, vanilla CBM (adapted to sparse annotation settings; see~\ref{sup_section_vanilla CBM} in the Supplementary Material), and existing AGBD map products. More importantly, it provides interpretable intermediate outputs that are useful in their own right and which can be used to provide corroboration and confidence in the AGBD predictions. These results illustrate how process guidance enhances both predictive performance and scientific trustworthiness.

\subsection{Related Work} \label{section_related work}

CBMs \cite{pmlr-v119-koh20a} are designed to improve interpretability by introducing a layer of human-defined concepts between the input and the prediction. CBM typically decomposes the mapping $f(x)$ into $f(x) = g(h(x))$, where $h(x)$ predicts intermediate concepts and $g(\cdot)$ maps these concepts to the final output. In the original formulation \cite{pmlr-v119-koh20a} the concepts are assumed to be independent and fully annotated, which provides a clear intervention point but can reduce performance in complex tasks \cite{pmlr-v119-koh20a}. To overcome these issues, several extensions have been proposed.

The risk of “concept leakage” in vanilla CBMs, where unintended information bypasses the bottleneck \cite{margeloiu2021concept}, motivated the introduction of hard CBMs which enforce discrete, binarised concept predictions to reduce leakage \cite{havasi2022addressing}. Probabilistic CBMs extend this idea further by parametrising concepts as distributions, allowing the model to express uncertainty in concept predictions and improve robustness under noisy labels \cite{kim2023probabilistic}.

Other variants of CBMs explore more flexible representations of concepts. Concept Embedding Models \cite{espinosa2022concept} represent each concept by a learnable embedding vector, from which probabilities are inferred, while stochastic or variational formulations allow richer modelling of dependencies \cite{vandenhirtz2024stochastic}. Multimodal models such as Contrastive Language-Image Pre-training \cite{radford2021learning} are used to automatically discover and assign concepts in order to reduce the dependence on fine-grained manual annotations \cite{yuksekgonul2022post, oikarinen2023label}, while \cite{laguna2024beyond} fine-tunes pre-trained black-box models on small validation sets with concept annotations, to increase their interpretability.

Together, these works demonstrate that CBMs have evolved into a diverse family of models balancing interpretability, performance, and flexibility. However, most CBM variants remain largely generic as they rely on predefined concepts or automatically discovered ones, but often lack grounding in domain-specific processes and casual dependencies. This limits their applicability in scientific domains where interpretability depends not only on human-readable concepts, but also on alignment with causal biophysical mechanisms.

In the proposed PG-CBM, intermediate concepts are biophysically meaningful attributes that are in themselves useful for users (e.g., tree canopy cover and height, tree stem number density). An aggregation module uses the intermediate attributes to produce the final output, according to known causal relationships. Unlike standard CBMs, where concepts are fixed, independent, and often generic, PG-CBM (i) embeds domain knowledge to guide the mapping between input data, intermediate concepts, and final output, (ii) allows flexible optimisation of the intermediate states during training, so they remain informative while still interpretable, and (iii) supports heterogeneous supervision by leveraging multiple data sources for different concepts. In this way, PG-CBM combines concept-based modelling with process-based reasoning, making it more suitable for scientific domains where trust, causality, and interpretability are critical.

\section{Use Case: Mapping Above Ground Biomass Density using Earth Observation Data}
\label{section_usecase}

To evaluate PG-CBM in a realistic scientific context, we apply it to the task of mapping AGBD from EO data. AGBD estimation represents a challenging, domain-specific problem where interpretability, domain-awareness, and robustness to label uncertainty and sparsity are essential. Direct measurement of AGBD involves harvesting, drying, and weighing tree mass, which is destructive and infeasible at scale \cite{gonzalez2018estimation}. Instead, forest inventory plots record more accessible tree attributes (such as tree diameter, canopy height, and the number of tree stems in a given area [stem number density]) with AGBD then being estimated indirectly using allometric equations that relate these variables to biomass \cite{gonzalez2018estimation, chave2004error}. However, these allometric equations are subject to significant site-level variability \cite{chave2014improved} and carry substantial and well-documented uncertainties \cite{chave2014improved, RejouMechain2014}.

Many EO data sources provide information related to AGBD, including active EO sensors such as SAR and Light Detection and Ranging (LiDAR), as well as passive optical and thermal observations. EO data have been widely used to estimate canopy-related characteristics, such as height and cover, which are strongly correlated with AGBD \cite{turton2022improving}. 

Several modelling approaches exist for AGBD estimation from EO data. Empirical models, such as statistical regressions and traditional Machine Learning (ML) methods like Generalised Additive Model (GAM), Random Forests, or Gradient Boosting \cite{schuh2020machine, mascaro2014tale}, while interpretable and straightforward to apply, often lack generalisability and struggle to capture complex ecological relationships, especially with sparse or noisy label data \cite{li2020high}. Physical and semi-empirical models, on the other hand, attempt to simulate the interaction between vegetation structure and electromagnetic signals using radiative transfer and vegetation models \cite{ulaby1990michigan, karam1992microwave, brolly2012matchstick}. Although grounded in theory and helpful for interpretation, these models require numerous input parameters that are rarely available at scale \cite{tian2023review}, and their structural simplifications relative to real-world complexity often increases the bias and error of the physical models \cite{jia2021physics, jiang2025jax}. These limitations highlight the need for more flexible, scalable, and interpretable approaches for AGBD estimation \cite{santoro2024biomasscci, chave2014improved, mitchard2009using}.

In recent years, DL has become a leading approach for estimating AGBD from EO data \cite{lang2023high, lang2022global, weber2025unified}, offering superior capacity for modelling complex, non-linear relationships in multi-source, spatio-temporal EO data, such as SAR, optical, and LiDAR. The availability of large-scale spaceborne LiDAR datasets, such as Global Ecosystem Dynamics Investigation (GEDI) \cite{dubayah2020global}, has further accelerated progress by providing canopy structure measurements suitable for DL training and validation. However, existing DL approaches face persistent challenges. Label data derived from LiDAR or field plots are often sparse, uncertain, and inconsistently calibrated across biomes, limiting model generalisability. Moreover, most DL models remain black-box systems with limited interpretability or grounding in biophysical processes, reducing their trustworthiness in scientific applications. To address these issues, PG-CBM mirrors the reasoning process commonly used in ecological practice, where AGBD is indirectly derived from measurable tree attributes. Inspired by this process, PG-CBM embeds ecological knowledge directly into the model architecture, encouraging the alignment of its predictions with underlying causal relationships, rather than relying solely on unconstrained statistical correlations. A further advantage is that the model can leverage label data corresponding to any of the intermediate attributes, substantially expanding the available label data and improving supervision efficiency.

\subsection{Dataset and Preprocessing} \label{section_dataset}

In this study, we focus on dry tropical woodlands and savannas of southern and central Africa, corresponding to the spatial extent of a unique large-scale corpus of field-plot data available from the SEOSAW network \cite{seosaw2021network} (Figure~\ref{sup_fig_SEOSAW_a} in the Supplementary Material). Within this domain, we use SAR and optical imagery as EO input features, along with longitude and latitude coordinates as positional data. For training labels, we use both i) sparse heterogeneous field data where AGBD and stem number are estimated for each plot \cite{seosaw2021network} and ii) widespread, numerous canopy height and cover metrics from the GEDI space-borne LiDAR (L2 footprint-level observations) \cite{dubayah2020global}. It should be noted that the two sources of the supervision labels (i.e., field-plots for stem number density and AGBD, and GEDI for canopy cover and height) differ widely in availability, with $\sim$8.7 million GEDI-labelled patches and 8,260 plot-labelled patches. For AGBD estimation, all available 1\,ha plots (162 plots) are used as the validation set (due to lower geolocation errors and edge effects \cite{RejouMechain2014}, as well as their spatial variability and independence from the training data; see Fig.~\ref{sup_fig_SEOSAW_a} in the Supplementary Material). The remaining plots of varying shapes and sizes are used for training (8,098 plots).

These datasets differ in spatial resolution, measurement characteristics, and acquisition frequency, so careful preprocessing, including spatial alignment, masking, and normalisation, is required to ensure consistency across modalities. More detailed explanation of the datasets is provided in ~\ref{sup_section_dataset} in the Supplementary Material. Fig.~\ref{sup_fig_SEOSAW} in the Supplementary Material shows the training (in white) and validation plots (in red and orange), and an example of the label data footprints.

\section{Methodology} \label{section_Methodology}

\subsection{Background: Concept Bottleneck Models (CBMs)} \label{section_Background: Concept Bottleneck Models (CBMs)}

Vanilla CBMs decompose the predictive function into two stages:
\begin{equation}
    f(x) = g\!\big(h(x)\big),
\end{equation}
where $x \in \mathcal{X}$ is the input (e.g., EO data), $h: \mathcal{X} \rightarrow \mathcal{Z}$ maps inputs to a vector of concept activations $\hat{z}=h(x)\in \mathcal{Z}$, and $g: \mathcal{Z} \rightarrow \mathcal{Y}$ maps these predicted concepts to the target output $\hat{y}$. 
The joint training objective is:
\begin{equation}
    \mathcal{L}_{CBM} = \lambda \, \mathcal{L}_{\text{concept}} \big(h(x), z \big)
     + (1 - \lambda) \, \mathcal{L}_{\text{task}} \big(g(h(x)), y \big),
\end{equation}
where $z$ are annotated concept labels (if available), $y$ is the final label, and $\lambda$ balances concept- and task-level supervision. 

CBMs provide interpretability by enforcing that prediction flows through human-understandable intermediate concepts. 
However, they make simplifying assumptions that limit their scientific applicability: (i) full concept supervision is available, (ii) concepts are independent, and (iii) the aggregation function $g(\cdot)$ is purely empirical, not grounded in known causal processes.

\subsection{Process-Guided Concept Bottleneck Model (PG-CBM)} 
\label{section_PG-CBM}

\begin{figure*} [!t]
	\centering
		\includegraphics[scale=.15]{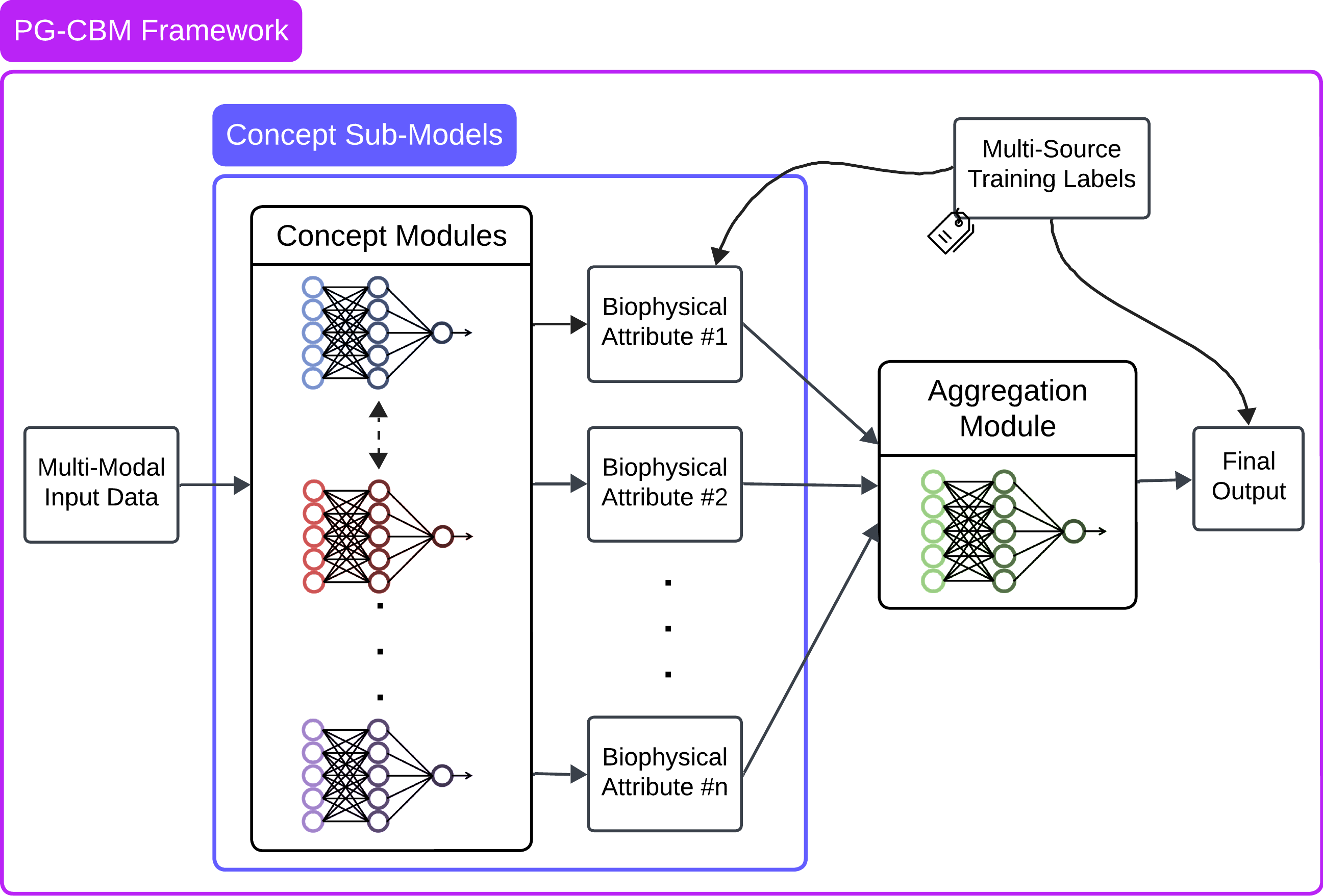}
	\caption{Concept map of the PG-CBM framework. The framework consists of concept and aggregation modules, each implemented as sub-models within the DL architecture. The concept modules predict intermediate biophysical attributes (i.e., ecologically meaningful features that serve as interpretable representations), while the aggregation module combines them to estimate the final target variable.}
	\label{fig_concept_map}
\end{figure*}

We extend CBMs into the Process-Guided CBM (PG-CBM), a framework that integrates domain-specific causal knowledge directly into the model structure.
PG-CBM decomposes the predictive mapping as:
\begin{equation}
    f(x) = g\big(h_{1}(x), h_{2}(x), \ldots, h_{k}(x)\big),
\end{equation}
where each sub-model $h_i: \mathcal{X} \rightarrow \mathcal{Z}_i$ estimates a domain-defined intermediate variable (e.g., canopy cover, canopy height, stem number density), and $g : \mathcal{Z}_{1} \times \cdots \times \mathcal{Z}_{k} \to \mathcal{Y}$ aggregates them to estimate the target (e.g., AGBD). 
Unlike vanilla CBMs, PG-CBM explicitly aligns $\mathcal{Z}_i$ with known causal mechanisms, reflecting the true process:
\begin{equation}
    x \;\xrightarrow{h_i}\; \hat{z}_i \;\xrightarrow{g}\; \hat{y}, 
    \qquad y \approx g(z_{1}, z_{2}, \ldots, z_{k}).
\end{equation}
This alignment ensures that the bottleneck variables correspond to biophysically meaningful quantities rather than arbitrary semantic tags. Fig.~\ref{fig_concept_map} shows the concept map of PG-CBM framework. 

\subsection{Theoretical Insight: Why Process Guidance Improves Generalisation}
\label{section_theoretical_insight}

PG-CBM introduces a structural inductive bias that constrains the learned mapping 
\begin{equation}
    f(x)=g(h_1(x),h_2(x),\ldots,h_k(x))
\end{equation}
to follow a causal process consistent with domain knowledge. Each sub-model $h_i$ represents a causal state variable, and $g(\cdot)$ approximates the biophysical mechanism linking them to the target variable.

\paragraph{Structural regularisation}
In unconstrained deep networks, the hypothesis space $\mathcal{F}_{DL}$ can represent any correlation present in the data.
PG-CBM restricts the function class
\begin{equation}
    \begin{aligned}
    \mathcal{F}_{PG-CBM}=&\{g\!\circ\!h_1,\ldots,h_k \,|\, (z_1,\ldots,z_k)\},
\end{aligned}
    \end{equation}
to follow the known process. Since $\mathcal{F}_{PG-CBM} \subset \mathcal{F}_{DL}$, the effective Rademacher complexity \cite{truong2022rademacher} satisfies $\mathcal{R}(\mathcal{F}_{PG-CBM}) < \mathcal{R}(\mathcal{F}_{DL})$, reducing model variance and discouraging spurious correlations. Therefore, the expected generalisation error bound is tighter for PG-CBM than for a purely black-box model;
\begin{equation}
    \begin{aligned}
    \mathbb{E}[|f(x)-y|] \le \text{empirical training error} + O(\mathcal{R}(\mathcal{F})).
\end{aligned}
    \end{equation}    

\paragraph{Causal invariance}
We assume that the true causal structure over $(X,Z,Y)$ satisfies $p(X,Z,Y)=p(X)p(Z\!\mid\!X)p(Y\!\mid\!Z)$ (i.e., causal intermediate variables $Z$ mediates the relationship between $X$ and $Y$). While black-box models approximate $p(Y\!\mid\!X)$ directly, PG-CBM explicitly learns $p(Z\!\mid\!X)$ and $p(Y\!\mid\!Z)$, enforcing the direction $X\!\rightarrow\!Z\!\rightarrow\!Y$. When the input space $p(X)$ is modified (due to sensor differences, environmental changes, or new geographic domains), but the underlying causal mechanisms are unchanged, the conditional relation $p(Y\!\mid\!Z)$ remains invariant \cite{scholkopf2021toward, ahuja2023interventional}, providing theoretical justification for PG-CBM’s improved OOD robustness.

Process guidance thus acts as a \textit{causality-preserving regulariser}, restricting the hypothesis space to functions that follow known mechanisms and encouraging representations invariant to irrelevant distributional changes.

\subsection{Theoretical Distinction from Vanilla CBMs}
\label{section_Theoretical Distinction from Vanilla CBMs}

Beyond outperforming black-box models, PG-CBM is a process-grounded generalisation of the vanilla CBM framework.

\paragraph{Process-grounded factorisation}
Vanilla CBMs factorise $f(x)=g(h(x))$ under an implicit assumption that the intermediate concepts are independent, while PG-CBM accounts for the correlation between intermediate concepts and grounds the decomposition in a domain process graph $G_P$:
\begin{equation}
    \begin{aligned}
p_{PG-CBM}(X,Z,Y)=&p(X)\!\prod_{i=1}^k p(Z_i\!\mid\!\mathrm{Pa}_{G_P}(Z_i))\\&\,p(Y\!\mid\!Z_1,\dots ,Z_k),
\end{aligned}
    \end{equation}
where $\mathrm{Pa}_{G_P}(Z_i)$ denotes the parent variables of $Z_i$ in the process graph $G_P$. However, the graph itself is not explicitly constructed. Each conditional term $p(Z_i\!\mid\!\mathrm{Pa}_{G_P}(Z_i))$ is parametrised by a learnable sub-model $h_i(\cdot)$, allowing the model to learn the dependency structure from data under process-guided constraints. This factorisation embeds known or learnable dependencies among intermediate variables and ensures that the bottleneck reflects causal, rather than purely semantic, relationships.

\paragraph{Heterogeneous supervision}
Vanilla CBMs require complete $(x,z,y)$ tuples for training (recent variants, however, have relaxed this constraint \cite{yuksekgonul2022post, oikarinen2023label}), while PG-CBM supports partial and heterogeneous supervision:
\begin{equation}
    \begin{aligned}
\mathcal{L}_{PG-CBM}
   =&\sum_i \alpha_i \mathcal{L}_i(h_i(x),z_i)
     +\\&\beta \mathcal{L}_y(g(h_1(x),\dots,h_k(x)),y).
\end{aligned}
    \end{equation}
This allows each sub-model $h_i$ to exploit its own dataset and transfers causal information across concept spaces through $g(\cdot)$.

\paragraph{Causal consistency}
PG-CBM explicitly parametrises $g(\cdot)$ to reflect known functional relationships among concepts (e.g., monotonic or non-linear process laws), encouraging the learned $p(Y|Z)$ to respect the causal semantics. Let $Z^*$ denote true process variables and $Z$ the learned bottleneck. PG-CBM enforces $I(Z;Z^*) \gg I(Z;Y|Z^*)$, where $I(\cdot;\cdot)$ denotes mutual information, quantifying how much one variable informs about another. The inequality indicates that the learned concepts $Z$ retain more information about the true process variables than about any residual correlations with the target $Y$, encouraging causal consistency.

\paragraph{Hypothesis-space hierarchy}
The inclusion relationships among the hypothesis spaces of the model classes can be expressed as:
\begin{equation}
    \mathcal{F}_{PG-CBM} \subset \mathcal{F}_{CBM} \subset \mathcal{F}_{DL},
\end{equation}
with corresponding Rademacher complexities
\begin{equation}
    \mathcal{R}(\mathcal{F}_{PG-CBM}) < \mathcal{R}(\mathcal{F}_{CBM}) < \mathcal{R}(\mathcal{F}_{DL}).
\end{equation}
PG-CBM therefore benefits from tighter generalisation bounds and greater causal fidelity than both black-box and vanilla CBMs.

Overall, PG-CBM transforms CBMs from interpretability-focused models into more causally consistent and process-aware predictors that balance transparency, physical plausibility, and generalisation to bridge the gap between empirical DL and mechanistic modelling.


\subsection{Ecological Intuition of the Model Design} \label{section_Intuition}

With respect to AGBD estimation, several studies have constrained statistical models with biophysical variables, often by estimating intermediate attributes (typically canopy height and occasionally canopy cover) before deriving AGBD \cite{duncanson2022aboveground, saatchi2011benchmark, santoro2024design, ma2023novel, weber2025unified, bretas2023canopy}. Moreover, some studies have highlighted stem number density as a critical factor influencing the relationship between SAR backscatter and AGBD \cite{brolly2012matchstick, carreirasdeterminants}. The decoupled design of these models and other modelling limitations (see~\ref{sup_section_use case} in the Supplementary Material) restrict their applicability and accuracy for AGBD estimation. Inspired by these studies, we incorporate three ecological attributes as intermediate representations (i.e., concepts): (1) canopy cover, (2) canopy height, and (3) stem number density. Each attribute is modelled by its corresponding sub-model, and then aggregated to derive AGBD.

While the predicted intermediate attributes are not direct field measurements, they can be interpreted as the model’s estimations of those physical quantities. The aggregation module subsequently integrates these predictions and their learned causal dependencies to estimate AGBD, in a manner analogous to how allometric equations combine measured tree attributes to derive biomass.

PG-CBM explicitly guides the DL model to derive AGBD through ecologically known relationships rather than giving it full flexibility to discover the data-driven optimum mapping. We are aware of the risk that these constraints may prevent the model from reaching the mathematically optimal (i.e., loss-minimising) mapping between inputs (EO data) and outputs (AGBD). However, the trade-off is that these biophysical constraints reduce the likelihood of spurious correlations and improve interpretability, ultimately providing a more trustworthy decision-making pathway (see~\ref{section_theoretical_insight}). More detailed ecological intuition is provided in~\ref{sup_section_Intuition} in the Supplementary Material. 

\subsection{Model Architecture} \label{section_Model Architecture}

In this study, we employ a unified DL architecture across all sub-models within the PG-CBM framework (Fig.~\ref{sup_fig_Architecture} in the Supplementary Material). Each sub-model processes multiple EO data sources through dedicated encoder branches tailored to the unique characteristics of each modality (e.g., SAR, optical, positional data). These encoders capture data source–specific features before fusion, allowing the model to retain complementary information across modalities. The fused latent representation is then processed by a spatial pyramid module to extract features at multiple spatial scales. A series of multi-head self-attention units subsequently refines these representations by modelling long-range dependencies and cross-modal interactions, while residual and normalisation layers enhance stability and generalisation.

The enriched features are decoded through a two-stage process comprising a dense residual decoder and a shallow convolutional head, progressively transforming abstract latent features into semantically meaningful outputs. The final quantile regression head predicts multiple percentiles (\nth{10}, \nth{25}, \nth{50}, \nth{75}, and \nth{90}) of the target variable, providing probabilistic estimates that capture predictive variability and reduce bias from regression-to-the-mean effects. Dropout, diverse normalisation techniques, and non-linear activations are integrated throughout to improve generalisability and robustness against data imbalance. A detailed description of the model architecture is provided in~\ref{sup_section_Model Architecture} in the Supplementary Material.

Ensemble-based methods are a common and reliable way to estimate model uncertainty in many scientific applications, but they are computationally expensive for large DL models. As an alternative, the quantile regression head provides an intuitive measure of prediction variability and stability. By predicting multiple quantiles of the target variable, the model produces a conditional distribution rather than a single estimate, allowing the spread between quantiles to indicate how stable or variable the predictions are across regions. While not a formal uncertainty quantification, this variability measure supports more transparent interpretation of results.

\subsection{Pre-Training and Post-Training} \label{section_Training and Fine-tuning}

The PG-CBM framework is trained in two stages: (1) pre-training of the ecological attribute sub-models, and (2) end-to-end post-training (fine-tuning) of the complete model with knowledge transfer across the sub-models. Unlike conventional CBMs, which require complete $(x,z,y)$ supervision for joint optimization, PG-CBM supports heterogeneous supervision where each sub-model $h_i$ is trained using its own available data source. Training labels are drawn from both field-measured plot data and GEDI L2 footprint products, corresponding to different ecological attributes. Because both sources are spatially sparse (GEDI at footprint level covering discontinuous measurement points, and field plots over limited plot extents; see Fig.~\ref{sup_fig_SEOSAW_b} in the Supplementary Material), a masked loss function is used so that loss terms are computed only over pixels with valid labels.

During the pre-training stage, each sub-model $h_i$ learns to predict one ecological attribute from its corresponding data subset, ensuring that intermediate concept representations capture meaningful biophysical relationships prior to integration into the full framework. The canopy cover and canopy height sub-models are trained on $\sim$8.7 million patches sparsely labelled with GEDI data, while the stem number density and AGBD sub-models are trained on 8,098 patches sparsely labelled with field-measured plot data. The strong imbalance in training set sizes is due to the much greater availability of GEDI data compared to field plots, and illustrates PG-CBM’s capacity to exploit diverse, non-overlapping data sources.

In the post-training stage, the full model is fine-tuned end-to-end using ground-estimated AGBD as the target. This stage enables the aggregation module $g(\cdot)$ to learn the functional relationships among the predicted intermediate attributes and the final output, while allowing backpropagation to refine the sub-models jointly. Crucially, this fine-tuning does not collapse the interpretability of intermediate representations: each sub-model retains its ecological semantics while benefiting from shared gradients that capture inter-concept dependencies. This process-guided joint optimization distinguishes PG-CBM from both vanilla CBMs, which assume independent, fixed concepts, and generic multi-branch architectures that lack process constraints and causal alignment.

\subsubsection{Loss Function} \label{section_Loss Function}

To achieve stable, unbiased, and ecologically meaningful predictions, we developed an enhanced quantile-based loss function that combines focal weighting with multiple regularisation terms. The core of the loss function is a focal quantile loss to give higher importance to difficult and extreme samples that are often under-represented in the training data. Additional regularisation terms address failure modes and issues we observed during development: monotonicity to maintain valid quantile ordering, spatial consistency to promote smooth but realistic spatial transitions, quantile consistency to preserve proportional spacing between quantile estimates, and adversarial regularisation to reduce bias across high- and low-density regions of the target distribution. Together, these components ensure that the model learns robust, well-calibrated, and spatially coherent predictions while maintaining interpretability and resilience to data imbalance. See~\ref{sup_section_Loss Function} in the Supplementary Material for more detailed explanation of the loss function.

\section{Experimental Results and Discussion} \label{section_Experimental Results and Discussion}

\subsection{Experimental Results} \label{section_Experimental Results}

To evaluate the AGBD estimation performance and assess the effectiveness of process guidance, we compare PG-CBM against four benchmarks: a vanilla CBM (adapted to sparse annotation settings; see~\ref{sup_section_vanilla CBM} in the Supplementary Material), an equivalent black-box DL model, and two existing large-scale biomass products (the European Space Agency (ESA) Climate Change Initiative (CCI) biomass map v5.01 \cite{santoro2024biomasscci} and the GEDI L4B product \cite{dubayah2022gedi}. Full implementation and experimental details are provided in~\ref{sup_section_Experimental Setup} in the Supplementary Material.

Beyond overall predictive accuracy, we perform a series of domain-specific analyses designed to test whether the ecological knowledge embedded in PG-CBM genuinely improves scientific validity and domain awareness. These analyses examine the behaviour of intermediate attributes, their consistency with known biophysical causal relationships, and their contribution to the final AGBD estimates. By aligning the evaluation with established ecological principles, we aim to demonstrate that PG-CBM not only achieves competitive predictive performance but also yields representations that reflect realistic ecological dynamics, mitigate spurious correlations, and promote domain-aware DL models.

\subsubsection{Comparison with Vanilla CBM} \label{section_Comparison with Vanilla CBM}

\begin{figure} [!t]
	\centering
		\includegraphics[scale=0.42]{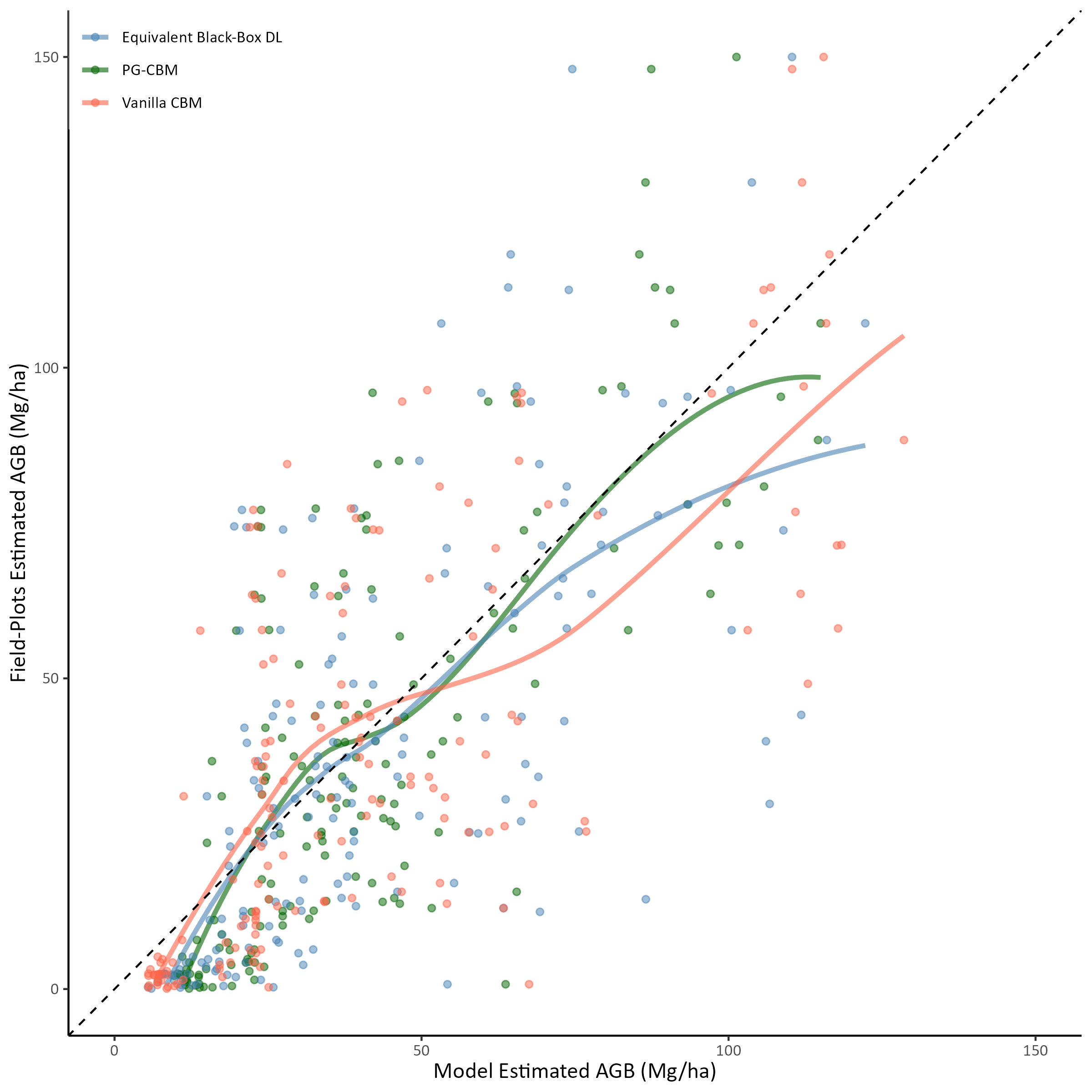}
	\caption{Comparison of the PG-CBM (green) with two benchmark DL models: the vanilla CBM (red; see~\ref{sup_section_vanilla CBM} in the Supplementary Material) and an equivalent black-box model (blue; see~\ref{sup_section_Equivalent black-box DL model} in the Supplementary Material). Each point represents ground- and model-estimated AGBD values, with a smooth trend line fitted for visualisation. The black dashed line indicates the 1:1 reference. Across the range of predicted AGBD values, PG-CBM exhibits a closer correspondence with the field estimates and reduced spread relative to the benchmarks, which show stronger deviations from the reference line, particularly at higher biomass.}
	\label{fig_Comparison_with_CBM and black-box}
\end{figure}

The vanilla CBM showed consistently worse performance than the PG-CBM (Fig.~\ref{fig_Comparison_with_CBM and black-box}; Table~\ref{table_qualitative_metrics}). Its  Root Mean Square Deviation (RMSD) (24.3\,Mg/ha) and mean bias (2.8\,Mg/ha) were both higher than those of PG-CBM (21.8\,Mg/ha and 1.5\,Mg/ha, respectively). Similarly, mean absolute bias and relative mean bias were larger for the vanilla CBM (18.6\,Mg/ha and 6.1\%) compared to PG-CBM (17.5\,Mg/ha and 3.2\%).

These differences reflect structural limitations of the vanilla CBM, where the concepts are predicted in a single bottleneck layer and passed unchanged to the final predictor. This works when concepts are fully observed and independent, but in practice, concepts are causally correlated and only sparsely supervised. PG-CBM addresses this by supporting heterogeneous supervision and by modelling them as interdependent ecological quantities. This design leads directly to the lower RMSD and reduced mean bias observed in Table~\ref{table_qualitative_metrics}.

\subsubsection{Comparison with Equivalent Black-Box DL Model} \label{section_Comparison with Baseline}

Degraded performance compared to the black-box models is a known problem with CBM models. However, the proposed PG-CBM outperforms the equivalent black-box DL model (Fig.~\ref{fig_Comparison_with_CBM and black-box}; Table~\ref{table_qualitative_metrics}). Specifically, PG-CBM reduces RMSD from 25.5\,Mg/ha to 21.8\,Mg/ha, halves the mean bias from 4.0\,Mg/ha to 1.5\,Mg/ha, and lowers relative mean bias from 8.7\% to 3.2\%, while increasing the interpretability and robustness.

This improvement results from two complementary factors. First, from a training data perspective, the black-box model learns only from plot-level AGBD values, whereas PG-CBM leverages additional intermediate attributes (GEDI canopy cover and height, and plot-based stem number density). This richer supervision signal stabilises training and yields more robust, less biased estimates. Second, from a domain-awareness perspective, PG-CBM embeds causal ecological structures in the learning process, while the black-box model, despite identical architecture, must approximate the complex EO–AGBD mapping without the domain guidance, making it more susceptible to overfitting and spurious correlations.

\subsubsection{Comparison with Existing AGBD Products} \label{section_Comparison with AGBD products}

\begin{figure} [!t]
	\centering
		\includegraphics[scale=0.42]{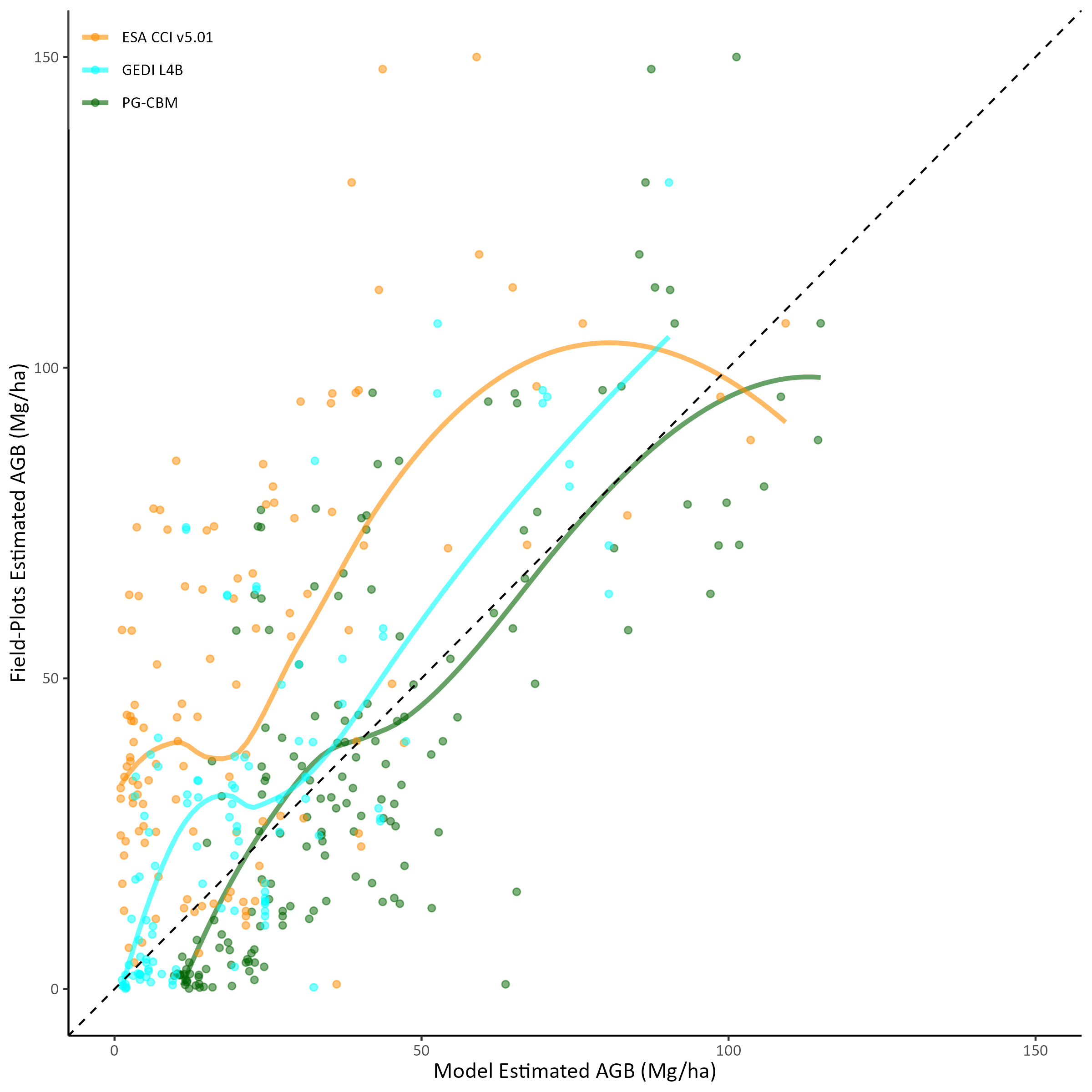}
	\caption{Comparison of the PG-CBM (green) with ESA CCI biomass map v5.01 (orange) and GEDI L4B product (cyan; see~\ref{sup_section_Other AGBD products} in the Supplementary Material). Each point represents ground- and model-estimated AGBD values, with a smooth trend line fitted for visualisation. The black dashed line indicates the 1:1 reference. Across the range of predicted AGBD values, PG-CBM exhibits a closer correspondence with the field estimates and reduced spread relative to the CCI and GEDI estimates, which show stronger deviations from the reference line at different biomass levels.}
	\label{fig_Comparison_with_CCI_GEDI}
\end{figure}

AGBD estimates from the proposed PG-CBM are both less biased and less variable than existing global AGBD products, such as ESA CCI biomass map and GEDI L4B product (Fig.~\ref{fig_Comparison_with_CCI_GEDI}, and Table~\ref{table_qualitative_metrics}). ESA CCI shows the weakest performance, with an RMSD of 38.5\,Mg/ha and a strong negative mean bias of -27.9\,Mg/ha (relative mean bias 60.4\%), highlighting systematic underestimation. GEDI L4B performs better, with the lowest RMSD (19.7\,Mg/ha) and absolute mean bias (13.9\,Mg/ha), but still shows a substantial negative mean bias of -8.0\,Mg/ha (relative mean bias 17.3\%). This shows high structure-dependent bias of GEDI L4B, discussed in~\ref{section_structure_dependent_bias}.

PG-CBM achieves a RMSD of 21.8\,Mg/ha, absolute mean bias of 17.5\,Mg/ha, and the lowest overall mean bias (1.5\,Mg/ha, 3.2\% relative mean bias). While GEDI L4B has slightly lower RMSD in low-biomass regions (visible in the lower left of Fig.~\ref{fig_Comparison_with_CCI_GEDI}), PG-CBM provides more reliable estimates in higher-biomass regions where GEDI underperforms. This distinction is critical because high-biomass areas dominate the carbon budget in large-scale ecological assessments. Hence, PG-CBM offers a stronger balance of accuracy and bias control, making it more suitable for regional and global AGBD mapping.

\begin{table}[!t]
    \centering
    \caption{Quantitative metrics comparing the AGBD estimates of different models.}
    \label{table_qualitative_metrics}
    \begin{tabular}{lccccc}
        \toprule
        \makecell{\textbf{Metric}\\\textbf{(Mg/ha)}}   
        & \makecell{\textbf{PG-CBM}}
        & \makecell{\textbf{Vanilla}\\\textbf{CBM}}
        & \makecell{\textbf{Black-box}\\\textbf{DL Model}}
        & \makecell{\textbf{ESA}\\\textbf{CCI}} 
        & \makecell{\textbf{GEDI}\\\textbf{L4B}} \\
        \midrule
        RMSD          & 21.8  & 24.3& 25.5   & 38.5  & 19.7 \\
        Bias          & 1.5 & 2.8& 4.0   & -27.9 & -8.0 \\
        Absolute Bias & 17.5  & 18.6&19.3    & 31.0  & 13.9 \\
        Relative Bias& 3.2\%& 6.1\%& 8.7\%& 60.4\%&17.3\%\\
        Interpretability& Moderate& Low& None&-&-\\
    \end{tabular}
\end{table}

\subsubsection{Structure Dependent Bias} \label{section_structure_dependent_bias}

An important challenge in AGBD mapping is structure-dependent bias, where estimation errors vary systematically with underlying vegetation structural properties, such as stem number density \cite{calders2015nondestructive}. Models that fail to account for these dependencies may perform well on average, but exhibit substantial over- or under-estimation in specific structural conditions. This not only reduces overall reliability but also undermines scientific and policy applications.

\begin{figure} [!t]
	\centering
		\includegraphics[scale=0.48]{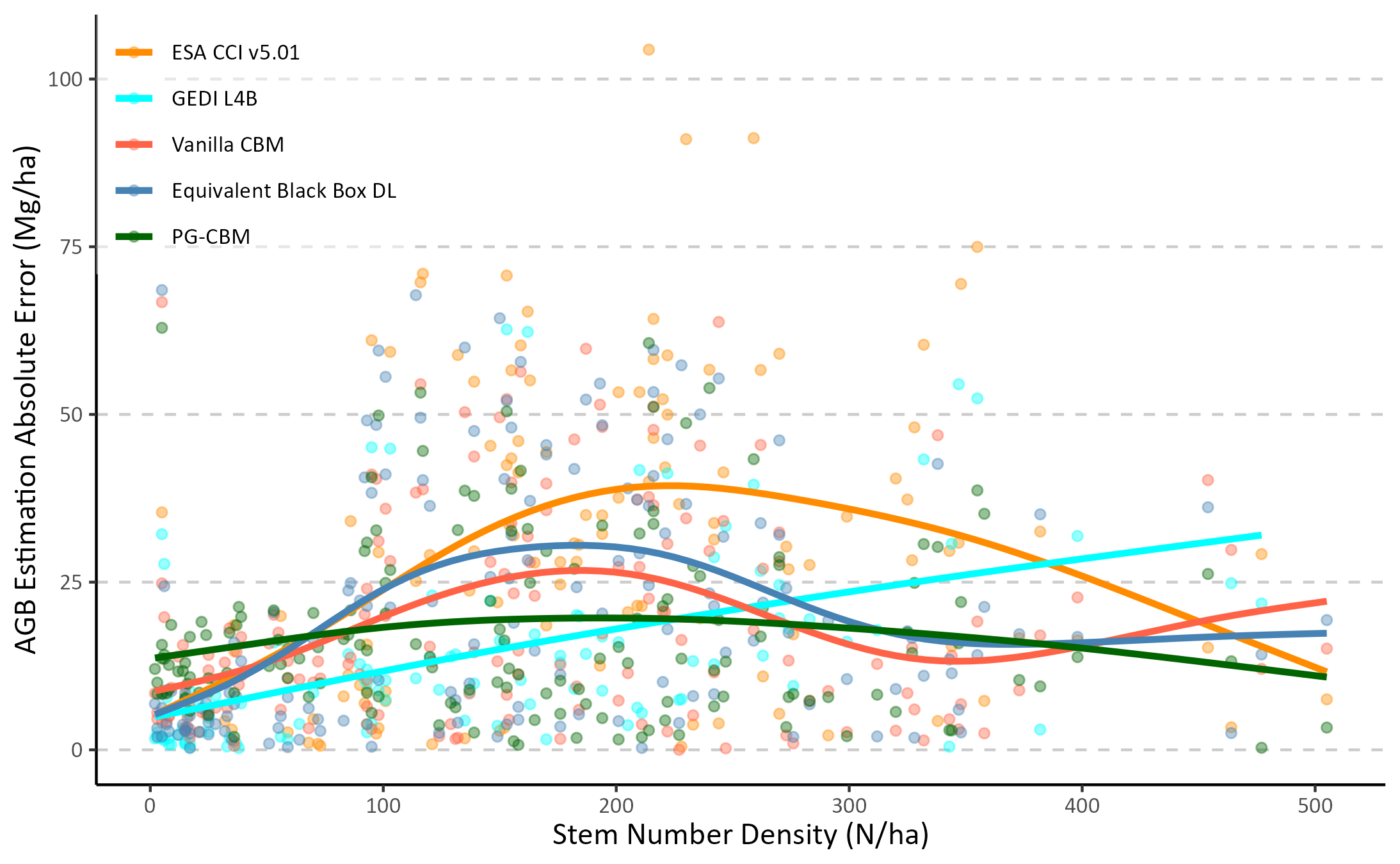}
	\caption{Comparison of AGBD estimation absolute errors across stem number density (N/ha) for PG-CBM, vanilla CBM, an equivalent black-box DL model, ESA CCI v5.01, and GEDI L4B. Each point represents the AGBD estimation error at a given stem density, with a smooth trend line. Unlike other models showing structure-dependent bias, PG-CBM maintains consistent accuracy across the full density range.}
	\label{fig_structure_dependent_bias}
\end{figure}

Fig.~\ref{fig_structure_dependent_bias} compares the prediction errors of PG-CBM against four benchmarks, across the stem number density gradient of the validation plots. All four benchmarks exhibit inconsistent estimation error for different vegetation structures and pronounced structure-dependent bias. CCI shows significantly higher estimation error at intermediate densities, while GEDI's prediction error increases at higher densities. Both vanilla CBM and equivalent black-box DL models show higher estimation error in lower intermediate stem number densities. In contrast, PG-CBM maintains a relatively flat error profile, with substantially reduced bias across the full density range. This consistency indicates that by embedding ecological knowledge into the DL model architecture, PG-CBM gains a better understanding of how tree attributes correlate with AGBD, avoiding spurious correlations that lead to bias in structurally heterogeneous conditions.

\subsubsection{Prediction Variability} \label{section_prediction_variability}

\begin{figure} [!t]
	\centering
		\includegraphics[scale=0.42]{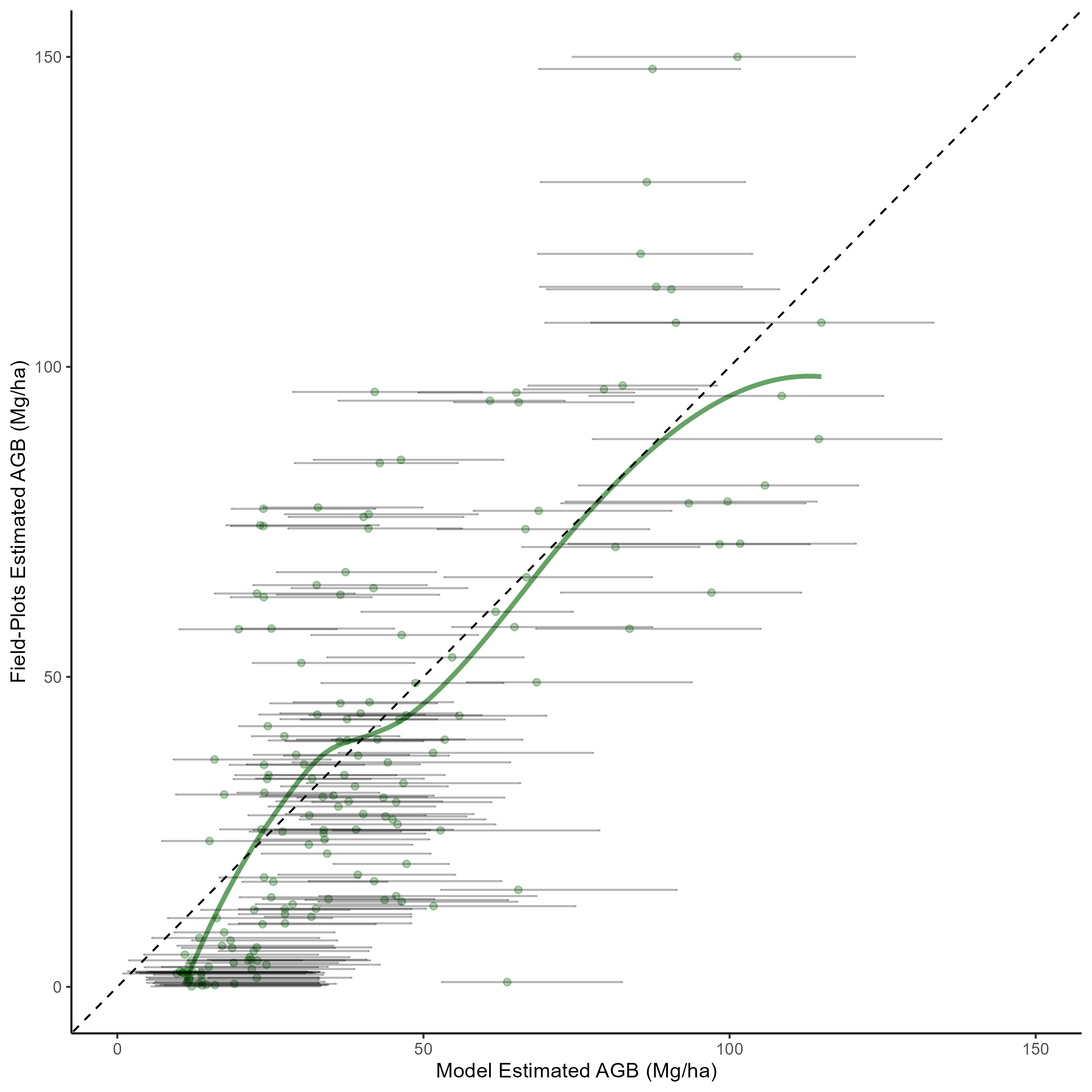}
	\caption{AGBD predictions from PG-CBM with prediction intervals. Green points show median predictions, grey bars denote the \nth{10}–\nth{90} percentile interval, the dashed line marks 1:1 agreement, and the green curve summarises the median trend. On average, PG-CBM prediction intervals span approximately $\sim$36±SD\% of the median estimate. Note that plot-derived AGBD values themselves also carry substantial uncertainty.}
	\label{fig_Error_Bars}
\end{figure}

Fig.~\ref{fig_Error_Bars} illustrates the prediction variability intervals of PG-CBM for each validation plot, defined as the range between the \nth{10} and \nth{90} percentile estimates from the model’s predictive distribution. The median prediction for each plot is shown as a green point, with the dashed line indicating 1:1 agreement with plot-estimated AGBD and the solid green curve summarising the overall trend across the dataset. These intervals reflect the spread of plausible model outputs given the input data and learned relationships.

Overall, the prediction intervals are well aligned with the variability observed in plot-estimated AGBD, providing an informative measure of model prediction stability. On average, the width of the \nth{10}–\nth{90} interval corresponds to $\sim$36±SD\% of the predicted median AGBD.

Importantly, prediction variability is not uniform across the biomass range. At low-to-moderate AGBD values, intervals are relatively narrow, reflecting the denser representation of training data in these ranges. At higher AGBD values, intervals widen, but the increase remains moderate, indicating that PG-CBM is more robust than black-box baselines that often exhibit unstable or unbounded variability in under-represented regions (see~\ref{section_Spurious Learning Resistance}). By explicitly modelling prediction intervals, PG-CBM provides both reliable point estimates and interpretable indicators of prediction stability, which are valuable for downstream applications such as carbon accounting and ecological monitoring.

\subsubsection{Causal Intercorrelation Between Intermediate Attributes} \label{section_Causal Intercorrelation Between Intermediate Attributes}

\begin{figure*} [!t]
	\centering
		\includegraphics[scale=0.8]{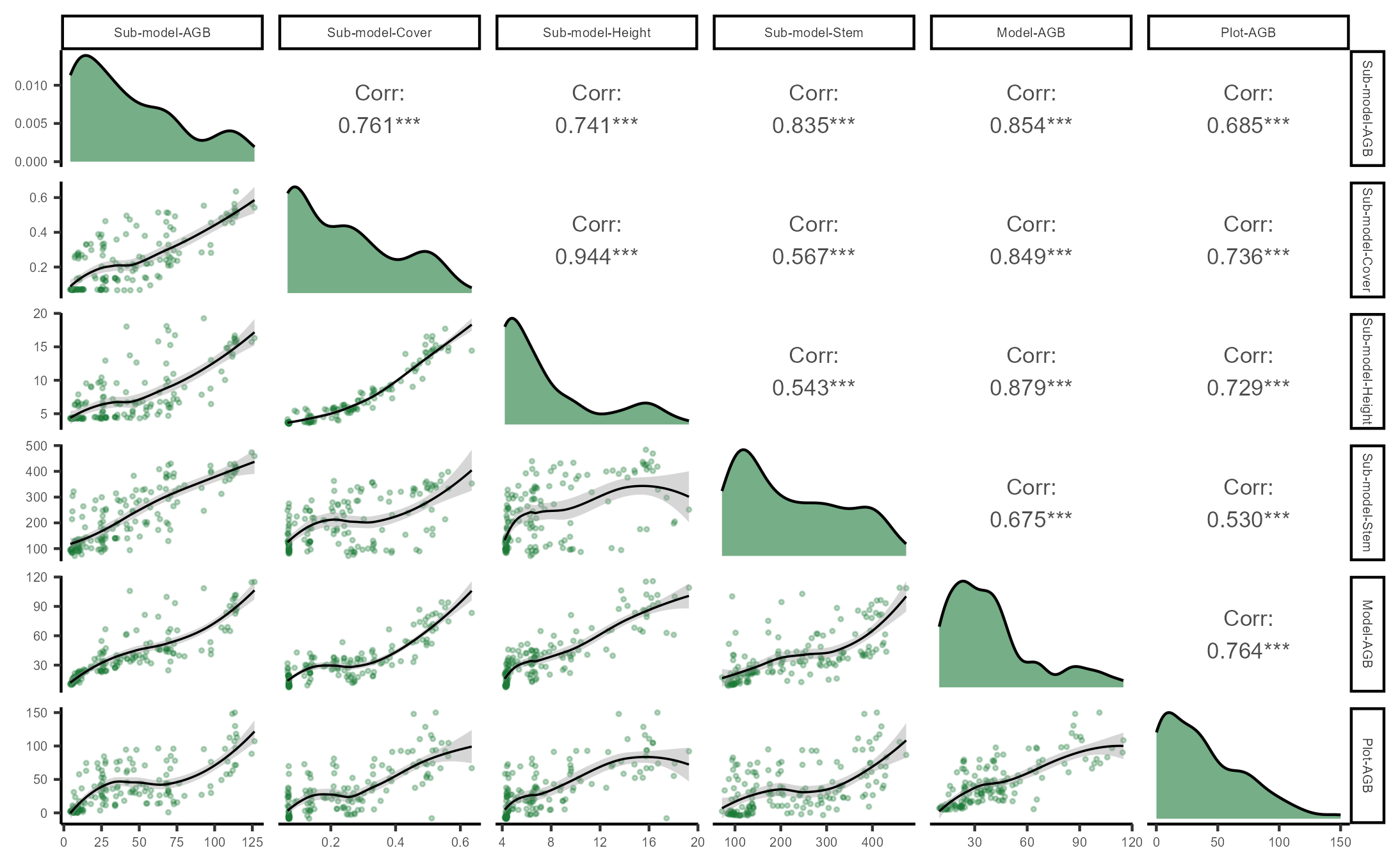}
	\caption{Pairwise relationships among intermediate attributes (canopy cover (\%), canopy height (m), and stem number density (N/ha)), final AGBD (Mg/ha) estimates from PG-CBM, and reference Plot-estimated AGBD (Mg/ha). Lower panels show scatterplots illustrating bivariate relationships, upper panels report corresponding Pearson correlation coefficients, and diagonal panels display individual variable distributions using kernel density estimates.}
	\label{fig_Sub_Model_Pairs}
\end{figure*}

PG-CBM models the causal interdependencies between the intermediate attributes through knowledge transfer between the sub-models with biophysically meaningful feature maps. These biophysical attributes are ecologically correlated and their latent representation can benefit from these interdependencies during post-training. There is a strong correlation ($\sim$0.94) between the estimated canopy cover and height (Fig.~\ref{fig_Sub_Model_Pairs}), highlighting both the reliable performance of the concept sub-models and their ability to capture causal relationships between intermediate biophysical attributes. 
Additionally, the scatterplots of the stem number density with canopy cover and canopy height show meaningful ecological patterns (Fig.~\ref{fig_Sub_Model_Pairs}). For instance, in areas with a high number of stems, the canopy cover tends to be denser, and trees often grow taller to compete for the sunlight, which is correctly represented in the intermediate representations. These ecological relationships evidence the ability of the proposed PG-CBM to capture domain-specific causal mechanisms.

\subsubsection{Interpretability and Domain-awareness Analysis} \label{section_Interpretability}

PG-CBM provides higher transparency into the DL model's internal reasoning. In Fig.~\ref{fig_Sub_Model_Pairs}, focusing on the PG-CBM AGBD estimates, we observe higher correlations with the modelled canopy cover ($\sim$0.85) and canopy height ($\sim$0.88) than with stem number density ($\sim$0.67). This suggests that canopy structure attributes have a higher impact on the DL model's decision-making pathway. This is a reasonable impact considering the dry tropical biome of the study area where tall high cover trees contribute significantly to the total biomass, compared to several small stems. This relationship is supported by the higher correlation of the plot AGBD with canopy cover ($\sim$0.74) and canopy height ($\sim$0.73) compared to stem number density ($\sim$0.53). The integrated higher explainability of PG-CBM and its ability to model domain-specific processes and causal relationships allow such conclusions and improve transparency and domain-awareness of DL models, necessary for scientific applications. 

\subsubsection{Spurious Learning Resistance Analysis} \label{section_Spurious Learning Resistance}

To evaluate the resistance of the proposed PG-CBM to spurious learning, we identified a subset of validation samples exhibiting OOD characteristics based on two criteria: (1) geographically remote plots that are poorly represented in the training data, and (2) plots isolated in the tree attribute–AGBD space, using stem number density as the representative attribute (other attributes originate from external sources (i.e., GEDI) and are unavailable for the same plots). Approximately one-third of the total validation samples were classified as OOD, while the remaining two-thirds served as In-Distribution (ID) samples (shown as orange and red points in Fig.~\ref{sup_fig_SEOSAW_a} in the Supplementary Material, respectively).

Figure~\ref{fig_spurious_comparison} compares the AGBD estimation absolute errors of the PG-CBM, vanilla CBM, and equivalent black-box DL models across ID and OOD sets. As expected, all models show reduced performance in OOD conditions. However, both concept-based models demonstrate better consistency and smaller degradation in predictive accuracy compared to the equivalent black-box model, with PG-CBM exhibiting the highest generalisability to OOD samples (mean bias of 7.2 Mg/ha and 9.1 Mg/ha versus 14.5 Mg/ha).

This robustness suggests that PG-CBM constrains the model to explore ecologically consistent solutions, therefore mitigating spurious correlations and overfitting to dataset-specific artefacts. In contrast, the unconstrained black-box model is more sensitive to OOD settings, resulting in larger errors and weaker generalisation to unseen regions. These findings confirm the theoretical conclusions from~\ref{section_theoretical_insight} and~\ref{section_Theoretical Distinction from Vanilla CBMs}.

\begin{figure} [!t]
	\centering
		\includegraphics[scale=0.35]{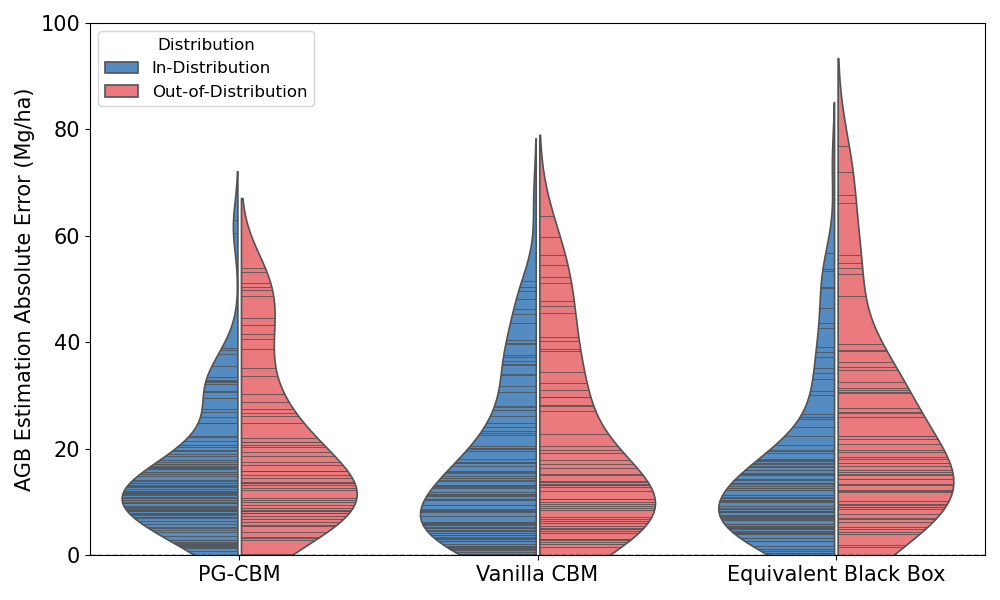}
	\caption{Comparison of absolute AGBD estimation errors for In-Distribution (ID) and Out-of-Distribution (OOD) samples across PG-CBM, vanilla CBM, and the equivalent black-box DL model. Each violin represents the distribution of absolute errors (Mg/ha), with inner sticks showing individual samples. Split violins highlight performance differences between ID (blue) and OOD (red) samples. PG-CBM exhibits the smallest performance degradation and reduced sensitivity to OOD samples, demonstrating stronger resistance to spurious learning than the baseline models.}
	\label{fig_spurious_comparison}
\end{figure}

\subsection{Beyond Accuracy: What does PG-CBM Offer?} \label{section_Beyond Accuracy: What PG-CBM Offers}

The proposed PG-CBM offers advantages that go well beyond reducing prediction errors. By embedding domain knowledge into the bottleneck and aligning intermediate attributes with established ecological causal processes, PG-CBM improves interpretability, flexibility, and robustness in a unified, domain-aware manner for scientific AI. These strengths address some of the challenges of applying DL to high-stakes domains, where sparse supervision, heterogeneous data, and the demand for physical consistency limit the usefulness of conventional black-box models.

\textbf{Process-guided concept representations:}
Unlike standard CBMs that rely on fixed, independent concept predictions, PG-CBM organises the bottleneck around process-guided intermediate attributes (e.g., canopy cover, canopy height, stem number density). These concepts are not only semantically interpretable but also causally linked to the final prediction via established ecological relationships. This preserves the improved interpretability while encouraging the model’s internal reasoning to align with the domain knowledge.

\textbf{Integration of multi-source and heterogeneous supervision:}
PG-CBM allows each intermediate attribute to be supervised independently, enabling training with multi-source data (e.g., field plots and GEDI footprints). This flexibility is particularly important in scientific applications (e.g., EO-based AGBD estimation), where labels for all attributes are rarely available for the same samples. By fusing heterogeneous supervision, PG-CBM leverages complementary information and improves robustness under sparse and noisy data conditions.

\textbf{Improved interpretability by design, not post-hoc:}
Instead of relying on post-hoc explainable AI (XAI) methods, which can be fragile or ambiguous, PG-CBM embeds explanation directly in its structure. Each sub-model predicts a biophysically meaningful attribute, and the aggregation stage combines them following ecological principles and causal relationships. This enables direct inspection of intermediate outputs, allowing domain experts to validate predictions and diagnose potential sources of error.

\textbf{Domain-aware learning and spurious resistance:}
By constraining the prediction pathway to pass through ecologically grounded intermediate concepts, PG-CBM reduces the risk of shortcut learning and spurious correlations. The model generalises more reliably to the OOD conditions because it is guided through pathways that are biophysically plausible, rather than arbitrary input–output mappings.

\textbf{Transparency and trust for deployment:}
In high-stakes applications such as climate monitoring and ecosystem management, trust is as important as accuracy. PG-CBM’s more transparent reasoning chain provides domain experts with clearer, and more verifiable checkpoints. The higher transparency is a step forward to build trust and make DL models more suitable for real-world scientific and policy contexts.

\textbf{Improved diagnosis and iteration:}
Because PG-CBM decomposes the prediction process, errors can be localised to specific intermediate attributes. This reduces the dimensionality of debugging and makes iterative improvements more targeted. For instance, if canopy height estimation is systematically biased, it can be detected and corrected independently without retraining the entire model.

\textbf{Enabling scientific insights:}
Beyond predictive accuracy, PG-CBM enables exploration of the relationships among intermediate attributes and their connection to input signals. This opens opportunities for generating new scientific insights, for example, by analysing how canopy structure affects SAR backscatter or how tree attributes impact biomass estimation (some examples shown in~\ref{section_structure_dependent_bias},~\ref{section_Causal Intercorrelation Between Intermediate Attributes}, and~\ref{section_Interpretability}). Thus, PG-CBM serves not only as a predictive tool but also as a framework for advancing scientific understanding.

\subsection{Limitations, Extensions, and Future Work} \label{section_limitations_future_work}

Overall, PG-CBM represents a step toward AI systems that not only predict but reason within the constraints of established scientific knowledge. While PG-CBM is demonstrated here for AGBD estimation, its underlying design principles extend broadly to scientific learning problems that require reasoning through interpretable intermediate states (e.g., \cite{worden2025combining}). For example, in other EO applications, PG-CBM could, in principle, be adapted to estimate soil moisture through variables such as surface temperature and precipitation, or to model carbon fluxes via leaf area index and photosynthetic capacity. In medical AI, biomarkers or disease progression stages could act as bottlenecks before diagnostic predictions, while in engineering, intermediate physical states such as stress or fluid-flow dynamics could precede performance outcomes. In this sense, PG-CBM extends the CBM family by embedding process-guided intermediate variables aligned with physical and ecological knowledge, offering a structured path to improve interpretability, robustness, and trustworthiness for deploying AI in high-stakes scientific contexts.

At the same time, several methodological limitations and corresponding opportunities for future research remain. First, PG-CBM relies on the availability of supervision for at least some intermediate variables (e.g., canopy height, cover, stem number density). While we show that combining heterogeneous supervision sources (e.g., GEDI LiDAR and field plots) is feasible, meeting this requirement might be harder in domains with limited measurements. Future research should explore weakly supervised learning, self-distillation from foundation models, and synthetic supervision from physics simulators to reduce reliance on direct labels.

Second, the current formulation embeds relatively simple ecological relationships, whereas real-world systems often involve hierarchical and dynamic dependencies such as demographic processes, disturbance events, or feedback loops. Integrating richer causal or dynamic priors through graph neural processes, differentiable simulators, or neural Ordinary Differential Equations (ODEs) could extend PG-CBM’s reasoning capacity beyond static structure–function mappings.

Third, while the framework performed consistently across the OOD validation set, its distributional robustness, domain transferability, and uncertainty propagation remain to be systematically tested. Promising avenues include Bayesian extensions of concept bottlenecks and ensemble-based uncertainty modelling.

Finally, significant advancements in the field depend on progress in multi-modal reasoning. Coupling PG-CBM with emerging vision–language and retrieval-augmented models could allow the fusion of EO imagery with textual ecological knowledge, plot metadata, or expert annotations, enriching both learning signals and interpretability. Such hybrid systems could serve as general-purpose scientific assistants that ground Large Language Models (LLMs) in measurable physical concepts for domain-aware prediction and explanation.

\section{Conclusion} \label{section_Conclusion}

In this study, we introduced the Process-Guided Concept Bottleneck Model (PG-CBM), an extension of the CBM family tailored for scientific applications where predictions are governed by intermediate biophysical attributes and domain-specific causal relationships. PG-CBM embeds domain knowledge directly into the architecture by constraining intermediate representations to align with biophysically meaningful quantities, which are then aggregated to produce the final prediction according to the known causal relationships. This mirrors the logic of many scientific workflows, where intermediate states mediate input–output relationships, and provides a mechanism for combining data-driven learning with domain-driven reasoning.

We evaluated PG-CBM through a case study on AGBD estimation from EO data. In this setting, the model predicts ecologically relevant attributes (canopy cover, canopy height, and stem number density) before aggregating them to estimate AGBD, reflecting how ecologists traditionally infer biomass through allometric scaling. This case study highlights both the potential and the challenges of applying PG-CBM in domains where interpretability, physical grounding, and robustness under sparse supervision are essential.

Our experiments show that PG-CBM reduces error and bias compared to black-box DL models, vanilla CBMs, and existing AGBD products, while producing interpretable intermediate outputs aligned with ecological causal relationships. These interpretable representations enable improved debugging, expose spurious learning, and open opportunities for extracting new scientific insights from the learned relationships.

Overall, the contribution of this work is demonstration of how extending CBMs with domain-specific process guidance can enhance interpretability and trustworthiness in data-driven models. PG-CBM represents a step toward bridging black-box DL with domain knowledge, offering a flexible yet interpretable alternative for scientific prediction tasks. As discussed in~\ref{section_limitations_future_work}, future work should further test its scalability, generalisation across domains, and integration with richer physics-based constraints, laying the groundwork for more robust and domain-aware AI systems.

\bibliographystyle{IEEEtran} 
\bibliography{refs} 

\vfill

\end{document}



\title{{\Large Supplementary Material:}\\[0.5em]
Process-Guided Concept Bottleneck Model}

\markboth{IEEE TRANSACTIONS ON PATTERN ANALYSIS AND MACHINE INTELLIGENCE, Vol.~V, No.~N, January~2026}%
{Shell \MakeLowercase{\textit{et al.}}: A Sample Article Using IEEEtran.cls for IEEE Journals}

\maketitle

\section{Use Case: Mapping Above Ground Biomass Density using Earth Observation Data} \label{sup_section_use case}

We use AGBD estimation from EO data as a use case to evaluate PG-CBM in a challenging, domain-specific setting. PG-CBM is intended as a step toward trustworthy AI systems in scientific applications, and AGBD estimation provides an example where explainability, domain-awareness, and reliable AI are critical. 

Consistent, timely and global data on AGBD is important for many environmental applications. The Global Climate Observing System (GCOS) has defined AGBD as an essential climate variable because of the role that biomass plays in mediating fluxes of energy and matter between the land and atmosphere. In addition, because biomass is $\sim$50\% carbon, accurate quantification of changes in AGBD are needed to quantify the global carbon budget, for carbon accounting and for international reporting to the UN Climate Change Convention. AGBD also provides information on forest productivity and other aspects of ecosystem functioning \cite{zemp2022gcos}. Direct measurement of above ground biomass involves harvesting, drying, and weighing tree mass and is destructive and infeasible at scale \cite{gonzalez2018estimation}. Instead, forest inventory plots record more accessible tree attributes (such as tree diameter, canopy height, and the number of tree stems in a given area [stem number density]) with AGBD then being estimated indirectly using allometric equations that relate these variables to biomass \cite{gonzalez2018estimation, chave2004error}. However, these allometric equations are subject to significant site-level variability \cite{chave2014improved} and carry substantial and well-documented uncertainties, particularly for large trees \cite{chave2014improved, RejouMechain2014}. These uncertainties propagate into field-based AGBD estimates that are used for model training, motivating the need for modelling approaches that can remain robust to noisy or uncertain supervision, such as the PG-CBM proposed here.

Forest inventory field measurements, while essential as reference data for AGBD estimation, are expensive to maintain, geographically biased toward accessible regions, and too sparse to capture the diversity in forest structure across large landscapes. Variability in measurement protocols, missed stems, and human error add further uncertainty, making high-quality, consistent AGBD data both scarce and heterogeneous \cite{chave2014improved, seosaw2021network, RejouMechain2014}. When combined with field measurements, EO offers a scalable and consistent approach for estimating AGBD across large areas.

\subsection{Earth Observation for AGBD Estimation} \label{sup_section_EO for AGBD}

Many EO data sources provide information related to AGBD. This includes active EO sensors such as Synthetic Aperture Radar (SAR) and Light Detection and Ranging (LiDAR) and passive observations of reflectance at optical and radiance at thermal and microwave wavelengths. EO observations have been widely used to estimate tree characteristics such as canopy height and cover, which correlate with AGBD \cite{turton2022improving}. SAR, especially lower frequency L- and P-band systems, is particularly useful due to its canopy penetration and strong correlation with AGBD \cite{quegan2019european, mcnicol2018carbon}. However, SAR is not a direct measurement of AGBD. Different vegetation attributes and soil conditions alter the relationship between radar backscatter and AGBD \cite{woodhouse2012radar}. Additionally, radar backscatter saturates at high biomass values (depending on the SAR frequency), resulting in a loss of sensitivity in some densely wooded ecosystems  \cite{joshi2017understanding}. These ambiguities combine with the limited and inconsistent label data available for calibration and validation \cite{meyer2022machine, Araza2022, RejouMechain2014} and the diversity of structure in the world's forests \cite{Mitchard2013, Zolkos2013} to pose severe constraints on reliable and unbiased large-scale AGBD estimations, even with EO.

Several modelling approaches exist for AGBD estimation from EO data. Empirical models, including statistical regressions and traditional Machine Learning (ML) methods like Generalised Additive Model (GAM), Random Forests, or Gradient Boosting, learn correlations between EO features and AGBD values \cite{schuh2020machine, mascaro2014tale}. While interpretable and easy to apply, they often lack generalisability and struggle to capture complex ecological relationships, especially with sparse or noisy label data \cite{li2020high}.

Physical and semi-empirical models aim to simulate the interaction between vegetation structure and electromagnetic signals using radiative transfer and vegetation models \cite{ulaby1990michigan, karam1992microwave, brolly2012matchstick}. Although grounded in theory and helpful for interpretation, these models require many input parameters that are rarely available at scale \cite{tian2023review}. Moreover, the reality of the vegetation structure and the physical interactions between the electromagnetic signal and the canopy are extremely complex and diverse, making it computationally and conceptually impractical for the physical models to replicate. Inevitably, a simplified model is used which increases the bias and error of the physical models \cite{jia2021physics, jiang2025jax}. These limitations highlight the need for more flexible, scalable, and interpretable approaches for AGBD estimation \cite{santoro2024biomasscci, chave2014improved, mitchard2009using}.

\subsection{Deep Learning for AGBD Estimation} \label{sup_section_DL for AGBD}

In recent years, DL has emerged as a powerful tool for EO tasks, including AGBD estimation. Compared to traditional methods, DL offers better modelling of complex, non-linear relationships in multi-source, spatio-temporal EO data, especially when integrating different EO data sources such as SAR, optical, and LiDAR. Its adaptability and superior performance for vegetation structure mapping have been demonstrated in multiple studies \cite{lang2023high, lang2022global, weber2025unified}. Progress in this field has relied heavily on the advent of spaceborne LiDAR, such as Global Ecosystem Dynamics Investigation (GEDI) \cite{dubayah2020global} and Ice, Cloud, and land Elevation Satellite 2 (ICESat-2) \cite{Markus2017}, which have provided a large dataset of canopy cover and height measurements, with sample size and coverage at a scale suitable for DL.

However, DL models for AGBD estimation come with their own challenges and limitations. A key limitation is the lack of label data. Whilst many studies have relied on estimates from spaceborne LiDAR systems as label data, these are not direct measurements, but instead are estimates based on biome-specific relationship between canopy height metrics measured by the LiDAR and AGBD, which have only been calibrated at a few locations. Field data present an alternative source of label data, but have not been available in sufficient quantity nor with the consistency required to train large DL models. Further challenges arise from lack of explainability and biophysics-awareness, which make the DL models more prone to spurious learning, hindering domain experts' trust in AI systems and their adoption in practical applications.

To address these issues, PG-CBM mirrors the reasoning commonly used in ecological practice, where AGBD is derived from tree attributes such as height, canopy cover, and tree stem number density, while also representing how SAR backscatter arises from these structural attributes. Inspired by this process, PG-CBM embeds this ecological knowledge directly into the model architecture, which predicts biophysically interpretable intermediate attributes, and subsequently uses these attributes to estimate AGBD. Thus the model’s reasoning pathway is structured around biophysical principles rather than relying solely on unconstrained statistical correlations. A further advantage is that the model can train on label data representing any of the intermediate attributes, greatly increasing the available training data corpus.

\subsection{Dataset and Preprocessing} \label{sup_section_dataset}

The multi-modal architecture (see~\ref{sup_section_Model Architecture}) enables us to integrate multiple sources of EO predictor data with both field-based and LiDAR measurements used to train and evaluate PG-CBM. In this study, we focus on dry tropical woodlands and savannas of southern and central Africa, corresponding to the spatial extent of a unique large scale corpus of plot data available from the SEOSAW network \cite{seosaw2021network} (Figure~\ref{sup_fig_SEOSAW_a}). Within this domain, we use SAR and optical imagery as EO input features, along with longitude and latitude coordinates as positional data. For training labels, we use both i) sparse heterogeneous field data where AGBD and stem number density are estimated for each plot and ii) widespread, numerous canopy height and cover metrics from the GEDI LiDAR (L2 footprint-level observations). These datasets differ in spatial resolution, measurement characteristics, and acquisition frequency, so careful preprocessing, including spatial alignment, masking, and normalisation, is required to ensure consistency across modalities.

\subsubsection{Input features} \label{sup_section_input features} \hfill\\
\textbf{SAR Data:} SAR systems record, \textit{inter alia}, the energy of an emitted radar signal that is scattered and returned to the sensor by the Earth surface, providing rich structural information for tree attribute mapping and AGBD estimation \cite{carreirasdeterminants, quegan2019european}. Lower-frequency SAR (e.g., L- and P-band) penetrates through the canopy and scatters from larger elements, including tree trunks, capturing critical information about AGBD. In this study, we use L-band SAR data from ALOS PALSAR-2, provided by the Japan Aerospace Exploration Agency (JAXA) \cite{JAXA_ALOS_PALSAR2}. We use the ScanSAR Level 2.2 product, which offers ortho-rectified and radiometrically terrain-corrected data with 25\,m spatial resolution. Data quality bitmask (MSK) is used to exclude corrupted pixels, and the annual median composite of the backscatter intensity ($\gamma^0$) for both horizontal-send, horizontal-receive (HH) and horizontal-send, vertical-receive (HV) polarisations, along with the local incidence angle (i.e., the angle formed by the radar irradiation direction and the normal of the ground slope), are used as SAR data inputs.

\textbf{Optical Data:} Passive optical sensors capture the reflectance of the sun's energy at multiple visible and infrared wavelengths, providing information about both biochemical and structural aspects of the soil and vegetation \cite{lang2023high}. To maximise the temporal frequency of observations, we use the Harmonised Landsat and Sentinel-2 (HLS) dataset, which creates a virtual constellation of optical satellites and provides global surface reflectance products at 30\,m resolution every 2-3 days \cite{Masek2021HLS}. In this work, we use the annual median composite of all 10 available spectral bands in the HLSL30 product as the source of optical data. Cloud, shadow, and snow/ice-affected pixels are masked out using the quality bitmask (Fmask).

\subsubsection{Training labels} \label{sup_section_ground truth} \hfill\\
\textbf{LiDAR Labels:} The GEDI LiDAR instrument, onboard the International Space Station (ISS), collects 3D vegetation structure data between 51.6$^\circ$N and 51.6$^\circ$S. GEDI L2 data provides footprint-level observations (25 m in diameter), spaced 60\,m along-track and 600\,m across-track \cite{dubayah2020global}, resulting in abundant measurements which cover about 4\% of Earth’s land surface over two years (\textgreater10 billion cloud-free footprints) \cite{dubayah2020global}. LiDAR systems emit laser pulses and record the return intensity as a function of time of flight/range, enabling near-direct measurements of canopy structure. As a result, derived metrics such as canopy cover and canopy height are considered near-direct observations, rather than modelled properties, making them well-suited as training labels. In contrast, the AGBD estimates from the GEDI L4B product are not direct measurements and if used as training labels, the modelling error (which is higher in regions such as the dry tropics due to sparse training data, see Fig.~\ref{fig_Comparison_with_CCI_GEDI}) will propagate through to the final estimation. In this study, we therefore do not use the AGBD estimates, but instead use canopy cover from the GEDI L2B product \cite{dubayah2020gediL2B} and canopy height (RH98, the relative height at which 98\% of energy has been returned) from the L2A product \cite{dubayah2020gediL2A} as training labels. Low-quality measurements are removed using the quality and geolocation flags, and footprints with low sensitivity (i.e., below the average canopy cover of all footprints with sensitivity $\geq$\,0.98) are excluded.

All EO datasets used in this study were preprocessed and downloaded using Google Earth Engine (GEE) \cite{gorelick2017google}. To ensure temporal consistency between EO observations and reference labels, the download periods were aligned with the timing of the plot or GEDI measurements: for samples labelled with GEDI observations, EO data were composited between August~2022 and August~2023; for samples labelled with plot data, EO imagery was extracted from the same calendar year as the corresponding plot census. All EO datasets were resampled and coregistered to the Coordinate Reference System (CRS) of the nearest ALOS scene. The Copernicus Global Land Cover map \cite{buchhorn2020cgls} was used to mask land covers that are not dominated by trees (i.e., urban/built up, bare/sparse vegetation, snow and ice, permanent water bodies, herbaceous wetlands, moss and lichen, and ocean/seas).

\begin{figure}[!t]
    \centering
    \subfloat[]{%
        \includegraphics[scale=0.22]{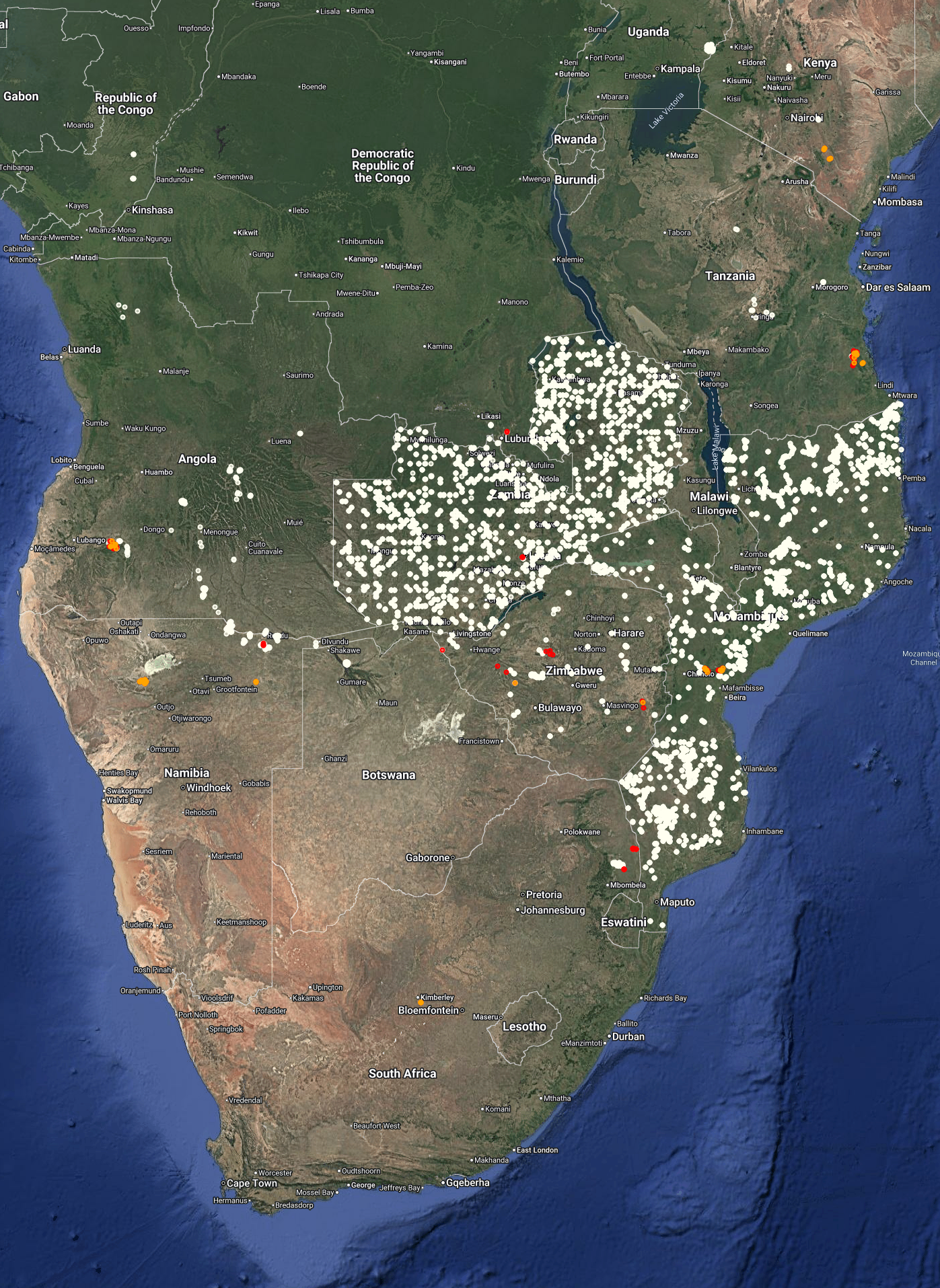}%
        \label{sup_fig_SEOSAW_a}
    }\\[0.1em]
    \subfloat[]{%
        \includegraphics[scale=0.2]{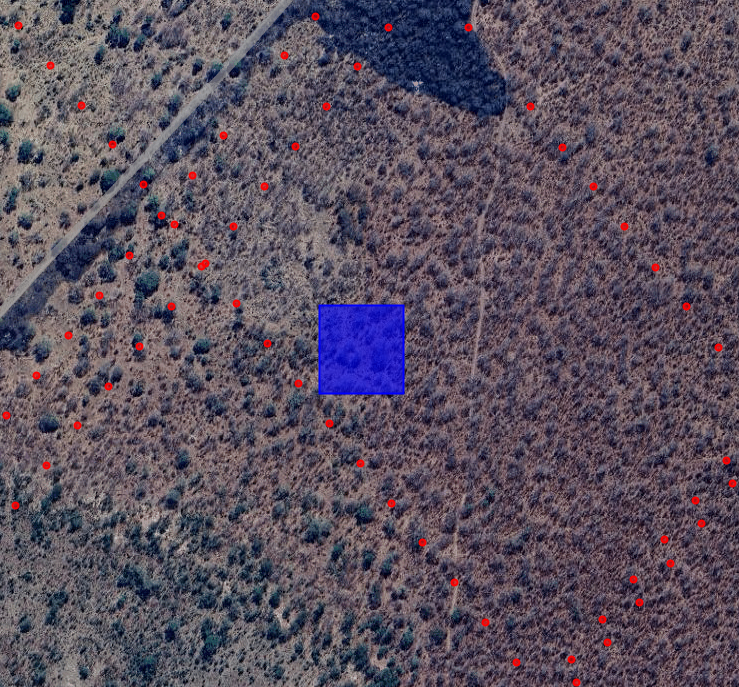}%
        \label{sup_fig_SEOSAW_b}
    }
    \caption{(a) Used SEOSAW plots: training in white, validation (all 1\,ha) in red and orange (orange points show the Out-of-Distribution (OOD) plots used in spurious learning analysis in~\ref{section_Spurious Learning Resistance}).
    (b) An example of label data showing a 1\,ha plot in blue and the available GEDI footprints in red.}
    \label{sup_fig_SEOSAW}
\end{figure}

\textbf{Field-Measured Plot Labels:} A field-measured plot refers to a designated area of land (plot) within a woody ecosystem where various vegetation attributes are systematically measured on the ground. These plots vary in size, shape, measurement techniques, and data collection standards. SEOSAW \cite{seosaw2021network} is a network of woodland survey plots across Africa, containing thousands of standardised plots with varying shapes and sizes. We use plot data from the SEOSAW network as label data and limit the spatial extent of the downloaded EO data to a 150\,km radius buffer around each plot's centroid. Plots with census dates after 2013 are selected to match EO data availability, and those lacking required measurements (i.e., stem number density, AGBD, and location data) are excluded. The number of tree stems per hectare (with diameter \textgreater10\,cm), along with their corresponding estimated AGBD per hectare (based on the allometry of \cite{chave2014improved}), are used as label data. This results in 8,260 plots of varying shapes and sizes, ranging from one-time circular plots with $\sim$9\,m radius ($\sim$0.025\,ha) to 25\,ha rectangular permanent plots. 1\,ha plots (162 in total) represent a "gold standard" (according to the Committee on Earth Observation Satellites (CEOS) biomass protocol \cite{duncanson2021aboveground}, SEOSAW \cite{seosaw2021network}) and are used as the validation set (due to lower geolocation errors and edge effects \cite{RejouMechain2014}, as well as their spatial variability and independence from the training data; see Fig.~\ref{sup_fig_SEOSAW_a}). The remaining plots of varying shapes and sizes are used for training (8,098 plots). Fig.~\ref{sup_fig_SEOSAW} shows the training (in white) and validation plots (in red and orange), and an example of the label data footprints. It should be noted that plot-estimated AGBD inherently carry large uncertainties, often with 95\% confidence intervals of 10–150\% of the mean, due to measurement errors, allometric model limitations, and sampling variability \cite{yanai2010estimating, clough2016quantifying}. 

\section{Ecological Intuition} \label{sup_section_Intuition}

With respect to AGBD estimation, several studies have attempted to constrain statistical models with biophysical variables, often by estimating intermediate attributes (typically canopy height and occasionally canopy cover) before deriving AGBD \cite{duncanson2022aboveground, saatchi2011benchmark, santoro2024design, ma2023novel, weber2025unified, bretas2023canopy}. However, in most of these approaches, the intermediate attributes are predicted separately and then used as fixed input features in the AGBD model. This decoupled design limits the model’s flexibility to jointly learn and optimise intermediate features that are most informative for biomass prediction.

Moreover, previous studies have highlighted stem number density as a critical factor influencing the relationship between SAR backscatter and AGBD \cite{brolly2012matchstick, carreirasdeterminants}. However, to the best of our knowledge, previous attempts to constrain statistical models with biophysical variables only considered canopy height and occasionally canopy cover. To address this issue, we incorporate three ecological attributes as intermediate representations (i.e., concepts): (1) canopy cover, (2) canopy height, and (3) stem number density. Each attribute is modelled by its corresponding sub-model, and then aggregated to derive AGBD.

\begin{figure*} [!t]
	\centering
		\includegraphics[scale=.15]{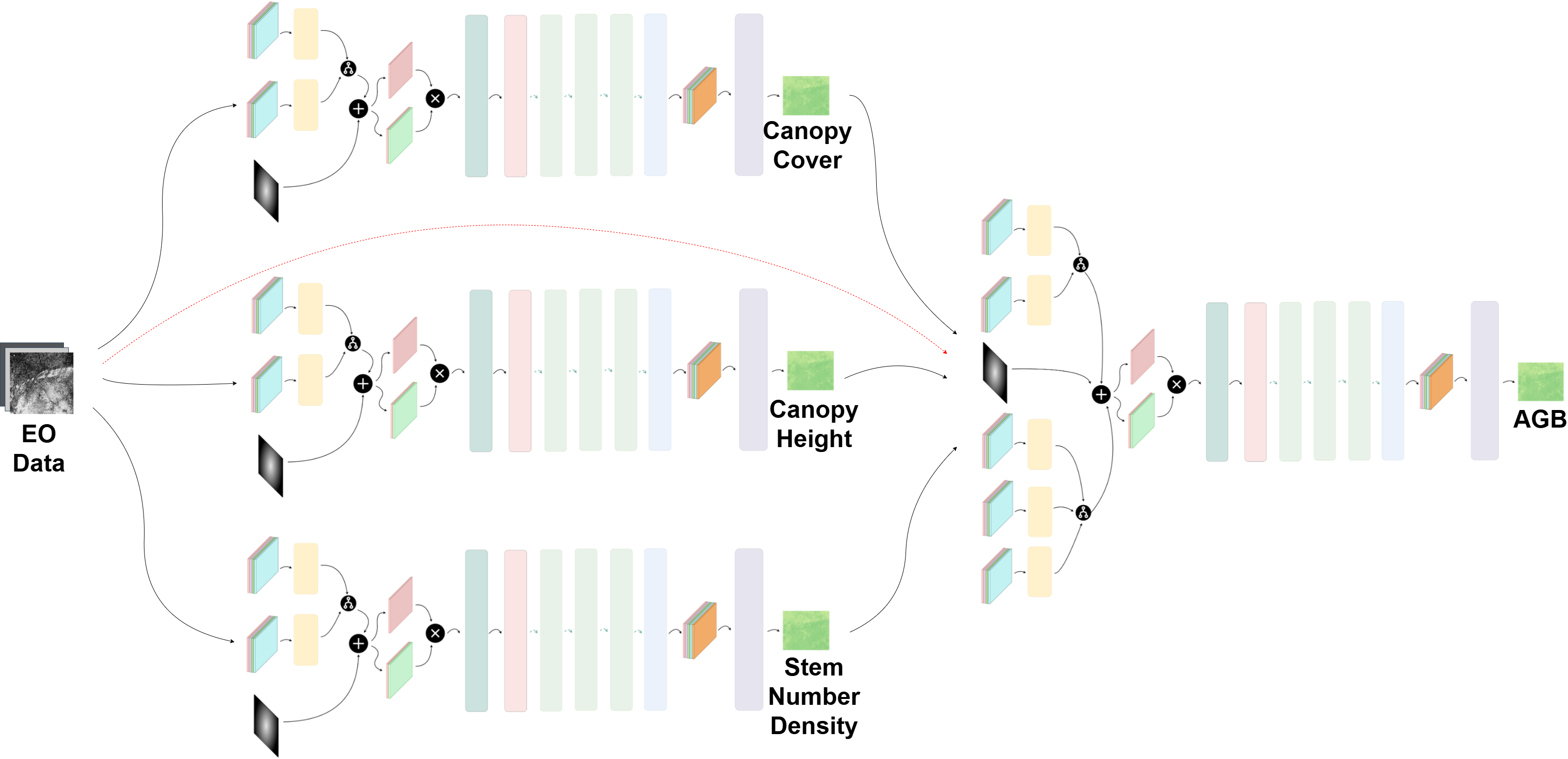}
	\caption{Schematic overview of the full PG-CBM framework, illustrating the ecological intuition behind its design. Intermediate feature maps correspond to biophysical attributes that serve as ecologically meaningful representations within the model. Full details of the DL architecture is provided in~\ref{sup_section_Model Architecture} and Fig.~\ref{sup_fig_Architecture}.}
	\label{sup_fig_Full_model}
\end{figure*}

For an ecologically intuitive understanding of PG-CBM and how it is designed to mirror the ecological process of AGBD estimation, consider the schematic view of the architecture with intermediate ecologically meaningful concepts in Fig.\ref{sup_fig_Full_model} (see~\ref{sup_section_Model Architecture} for details of the DL architecture and the sub-model modules). In a conventional black-box DL model, all of the intermediate feature maps are semantically meaningless (i.e., unexplainable). In contrast, PG-CBM assigns ecological meaning to selected intermediate feature maps and trains the model to estimate the corresponding tree attributes. Each of these modules function as a sub-model within the overall framework.

Thus, the DL model learns to estimate canopy cover, canopy height, and stem number density as intermediate attributes. While these estimates are not exact field measurements, they can be regarded as the model’s approximation of what would otherwise be measured in the field. The rest of the network then treats these semantically meaningful representations and other latent features as inputs to the next layer and combines them to estimate AGBD, analogous to how allometric equations use measured attributes to derive biomass.

PG-CBM explicitly guides the DL model to derive AGBD through ecologically known causal relationships rather than giving it full flexibility to discover the data-driven optimum mapping. We are aware of the risk that the constraints may prevent the model from reaching the mathematically optimal (i.e., loss-minimising) mapping between inputs (EO data) and outputs (AGBD). However, the trade-off is that these biophysical constraints reduce the likelihood of spurious correlations and improve interpretability, ultimately providing a more trustworthy decision-making pathway.

\section{Model Architecture} \label{sup_section_Model Architecture}

In this study we design a new DL architecture and use copies of the same architecture for each sub-model within the full framework. The new architecture consists of data source (modality)-specific encoder branches, spatial pyramid features, self-attention units, a two stage decoder block, and quantile regression heads. Fig.~\ref{sup_fig_Architecture} shows the designed architecture. For the purpose of reproducibility, we provide the detailed explanation of the architecture below:

\begin{figure*} [!t]
	\centering
		\includegraphics[scale=.085]{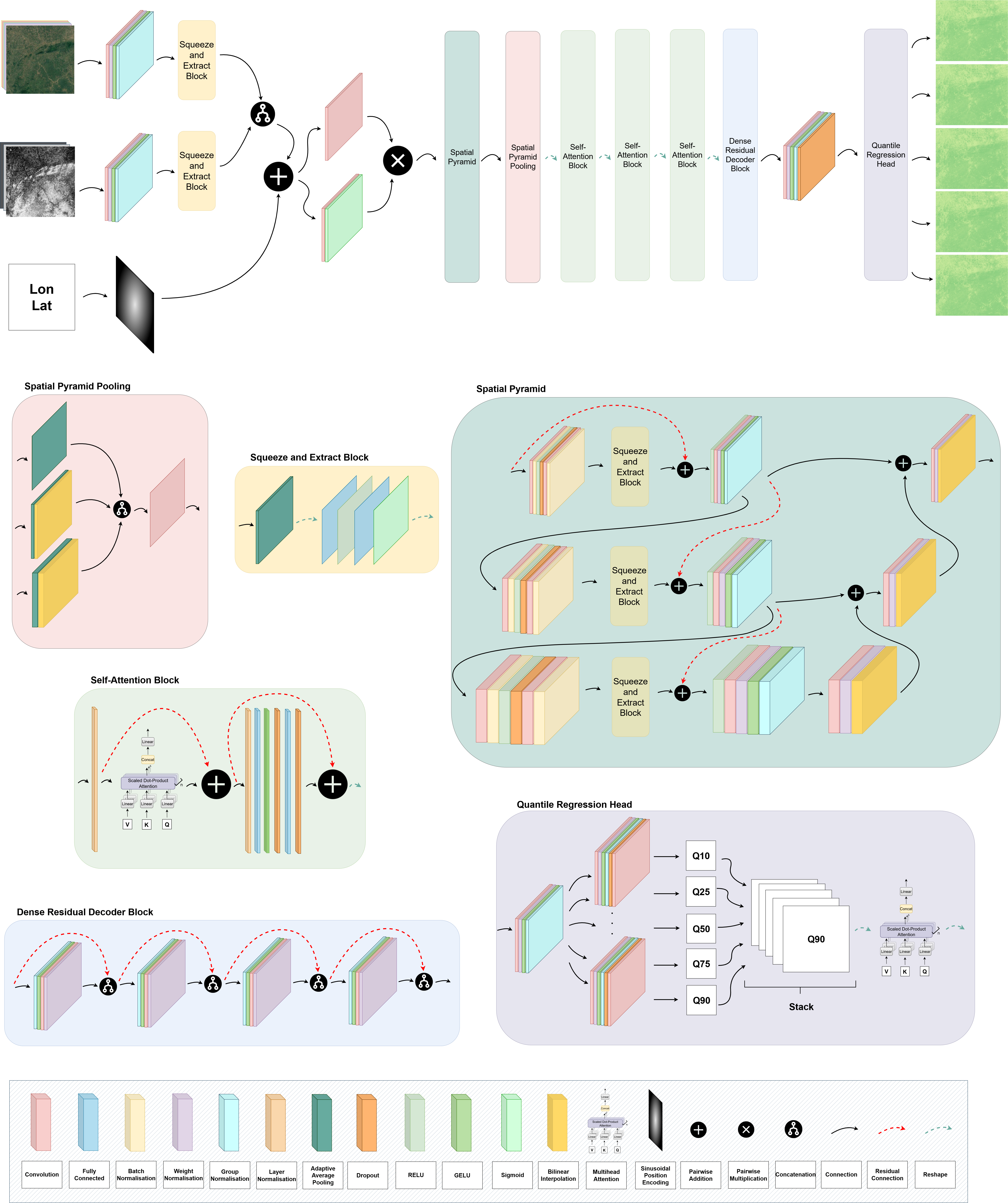}
	\caption{Architecture of the PG-CBM sub-model modules, consisting of data source (modality)-specific encoder branches, spatial pyramid features, self-attention units, a two stage decoder block, and quantile regression heads.}
	\label{sup_fig_Architecture}
\end{figure*}

\textbf{Data source-specific encoder branch:} Different EO data (e.g., SAR, optical) provide complementary but fundamentally different information about the target, with different statistical characteristics, noise patterns and true spatial resolutions (even at the same pixel spacing). A shared encoder would force different modalities into a premature single latent representation, which can lead to suboptimal feature learning and loss of data source-specific information. By processing each data source with a specific encoder branch, the model can learn optimal feature representation for each modality before combining the learned latent features \cite{ngiam2011multimodal}. This design also increases the flexibility of the architecture for removing or adding new data sources. As shown in Fig.~\ref{sup_fig_Architecture}, the encoder branch for EO data is a combination of convolution, fully connected, normalisation and pooling layers with different activation functions. Sinusoidal position encoding \cite{vaswani2017attention} is used as the encoder branch for the positional input data (i.e., longitude and latitude).

\textbf{Spatial pyramid:} The latent features extracted from each data source are fused together and processed with a spatial pyramid module for deeper feature extraction at multiple spatial scales, due to the multi-scale nature of AGBD features (e.g., small-scale such as gaps between trees in savannah, or large-scale such as landscape-scale patterns). The spatial pyramid module enables the model to learn both fine-grained and coarse contextual information by analysing features at different scales. A spatial pyramid pooling is used to aggregate the multi-scale features after the spatial pyramid module.

\textbf{Self-attention units:} Following the spatial pyramid feature extraction and pooling, the aggregated features are passed through three consecutive multi-head self-attention units to focus on different spatial patterns and cross-modal interactions simultaneously. The attention mechanism \cite{vaswani2017attention} enables the model to learn both short-range and long-range dependencies in the data and capture complex interactions between spatial features. Additionally, layer normalisation, fully connected layers, Gaussian Error Linear Units (GELU) activation functions, and residual connections are included in the self-attention modules to improve training stability, add smooth non-linearity and expand the capacity of the model.

\textbf{Decoder block:} After the self-attention modules, the enriched latent features are processed through a two stage decoder, consisting of a dense residual decoder block and a shallow convolutional decoder head. The dense residual decoder consists of a series of convolutional layers with group normalisation, GELU activation, and weight normalisation. Residual connections are used to facilitate gradient flow and preserve features from earlier stages of the block. After the dense residual encoder, the refined features are passed through a shallow convolutional encoder head with weight normalisation, GELU activation, and group normalisation. The second stage of the decoder is designed to further downsample and regularise the latent feature space for the final regression. These decoding layers play a critical role in bridging the gap between the abstract, high-dimensional latent feature space and the final semantically meaningful predictions (i.e., tree attributes and AGBD).

\textbf{Quantile regression head:} At the final prediction stage, we employ a quantile regression head for probabilistic prediction (\nth{10}, \nth{25}, \nth{50}, \nth{75}, and \nth{90} percentiles) of the target variable. It consists of convolutional layers with weight normalisation, GELU activation, group normalisation, and multi-head attention. Quantile regression provides probabilistic outputs and enables the model to estimate the predictive distribution of the target variable, a more informative output than a single-point estimate. It is particularly useful for regions with noisy or ambiguous EO data (e.g., saturation of SAR backscatter in dense biomass areas). Additionally, training with multiple quantile losses reduces prediction bias and prevents the model from consistently under- or over-estimating the target variable towards the mean (i.e., regression to mean bias).

Dropout is used throughout the architecture to improve the model's generalisability and reduce overfitting. It is especially important due to the imbalance and inconsistent training data. To ensure stable and efficient training, various normalisation layers, such as batch normalisation, layer normalisation, group normalisation and weight normalisation, are used in each module. Non-linear activation functions, such as Rectified Linear Unit (ReLU), GELU, and Sigmoid, are employed across the network to introduce non-linearity and enable the model to learn complex feature representations. Residual connections are used to facilitate smooth gradient flow and prevent information loss, particularly in the earlier stages of the model.

\subsection{Model Prediction Variability} \label{sup_section_Model prediction variability}

Ensemble-based methods are a common and reliable way to estimate model uncertainty in many scientific applications, but they are computationally expensive for large DL models. As an alternative, quantile regression provides a powerful and interpretable way to characterise prediction variability. Instead of producing a single point estimate, it outputs different quantiles (i.e., percentiles) of the target variable, which define a conditional predictive distribution. From these quantiles, one can construct prediction intervals that represent the spread of plausible outcomes given the input data. This approach is particularly relevant for AGBD estimation, where noise, ambiguity, and regional variability are inherent in both EO inputs and field data.

Prediction-interval maps derived from quantile regression highlight regions where model outputs are more variable, thereby flagging areas where estimates are less stable. While these intervals do not represent formal statistical uncertainty (e.g., confidence or credible intervals), they provide valuable information on the distribution of prediction and their robustness. Such indicators are especially useful for downstream applications like carbon accounting and forest management, and they align with best practices such as the CEOS biomass protocol \cite{duncanson2021aboveground}, which emphasises the importance of transparent variability assessment for EO-based AGBD products.

\section{Loss Function} \label{sup_section_Loss Function}

To ensure robust and unbiased estimation of the intermediate attributes, and the balance between overall prediction performance and preserving ecologically meaningful relationships, we developed an enhanced quantile loss function that combines multiple regularisation terms with focal weighting. The loss function is composed of five components:

\textbf{Focal Quantile Loss:}

The core component is a focal quantile loss that addresses the class imbalance by emphasising learning from more challenging samples. For a set of quantiles $Q = \{q_1, q_2, ..., q_K\}$, the focal quantile loss is defined as:

\begin{equation}
    \mathcal{L}_{quantile} = \frac{1}{K} \sum_{k=1}^{K} \frac{1}{N} \sum_{i=1}^{N} w_i^{(k)} \cdot \rho_{q_k}(y_i - \hat{y}_i^{(k)}),
\end{equation}
where $y_i$ is the training label, $\hat{y}_i^{(k)}$ is the $k^{th}$ quantile prediction, and $\rho_{q_k}$ is the standard quantile loss function:

\begin{equation}
    \rho_{q_k}(u) = \max(q_k \cdot u, (q_k - 1) \cdot u).
\end{equation}

The focal weighting mechanism $w_i^{(k)}$ combines error magnitude weighting with extreme value emphasis:

\begin{equation}
    w_i^{(k)} = (|y_i - \hat{y}_i^{(k)}| + \epsilon)^{\gamma} \cdot (1 + z_i^2),
\end{equation}
where $\gamma = 2$ is the focal parameter, $\epsilon = 10^{-6}$ prevents numerical instability, and $z_i$ is the z-score of the target value calculated using the mean $\mu_y$ and standard deviation $\sigma_y$ of the target values:

\begin{equation}
    z_i = \frac{|y_i - \mu_y|}{\sigma_y + \epsilon}.
\end{equation}

This weighting strategy is designed (1) to increase the focus on pixels with larger prediction errors, similar to focal loss in object detection, and (2) to give a higher importance to extreme values that are often under-represented in the training data but critical for accurate estimation.

\textbf{Monotonicity Regularization:}

In order to ensure valid probability distributions, quantile predictions must satisfy the monotonicity constraint $\hat{y}^{(1)} \leq \hat{y}^{(2)} \leq ... \leq \hat{y}^{(K)}$. We use the monotonicity regularisation term to penalise violations of this constraint:

\begin{equation}
    \begin{aligned}
        \mathcal{L}_{mono} =& \frac{1}{K-1} \sum_{k=1}^{K-1} \frac{1}{HW} \\& \sum_{h=1}^{H} \sum_{w=1}^{W} \text{ReLU}(\hat{y}_{h,w}^{(k)} - \hat{y}_{h,w}^{(k+1)}),
    \end{aligned}
\end{equation}
where $H$ and $W$ are the spatial dimensions. We employ ReLU to ensure that only positive monotonicity violations (i.e., where a lower quantile exceeds a higher quantile) are penalised. This regularisation term is essential for maintaining the probabilistic interpretation of the quantile outputs.

\textbf{Spatial Consistency Loss:}

To encourage spatial coherence in predictions without over-smoothing sharp edges and boundaries, we use an edge-preserving spatial regularisation term based on the Huber loss applied to gradients of the median quantile (i.e., \nth{50} percentile) predictions:

\begin{align}
    \mathcal{L}_{spatial} =
    &\tfrac{1}{2} \Bigg[ 
    \frac{1}{(H-1)W} \sum_{h=1}^{H-1} \sum_{w=1}^{W} 
      \sqrt{\epsilon + \big(\hat{y}_{h+1,w}^{(med)} - \hat{y}_{h,w}^{(med)}\big)^2} \nonumber \\
    &+ \frac{1}{H(W-1)} \sum_{h=1}^{H} \sum_{w=1}^{W-1} 
      \sqrt{\epsilon + \big(\hat{y}_{h,w+1}^{(med)} - \hat{y}_{h,w}^{(med)}\big)^2} 
    \Bigg],
\end{align}

where $\hat{y}^{(med)}$ is the median quantile prediction. This term helps to reduce the noise for smooth transitions between neighbouring pixels while avoiding over-smoothing to preserve ecologically meaningful transitions such as forest edges or clearings.

\textbf{Quantiles' Consistency Loss:}

To ensure that the spacing between adjacent quantiles reflects their theoretical relationship, we introduce a consistency loss that enforces proportional differences between quantile predictions:

\begin{align}
\mathcal{L}_{consistency} 
  &= \tfrac{1}{K-1} \sum_{k=1}^{K-1} 
     \mathcal{L}_{\text{SmoothL1}}\Bigg(
        \frac{\big|\hat{y}^{(k+1)} - \hat{y}^{(k)}\big|}{R_{pred}}, \nonumber \\
  &\hspace{3.5cm} q_{k+1} - q_k
     \Bigg),
\end{align}
where $R_{pred} = \max(\hat{y}^{(K)}) - \min(\hat{y}^{(1)}) + \epsilon$ is the prediction range used for normalization, and $\mathcal{L}_{SmoothL1}$ is the smooth L1 loss function:

\begin{equation}
    \mathcal{L}_{\text{SmoothL1}}(x, y) = 
    \begin{cases}
        0.5 (x - y)^2, & \text{if } |x - y| < 1 \\
        |x - y| - 0.5, & \text{otherwise}.
    \end{cases}
\end{equation}

The consistency term ensures that the gaps between the estimated quantiles are proportional to the theoretical quantile differences. This consistency is helpful for a more calibrated prediction variability.

\textbf{Adversarial Regularization:}

To reduce bias in predictions across different regions of the target distribution, we use a distribution-matching regularisation term based on the Jensen-Shannon divergence \cite{menendez1997jensen} between predictions in high-density (i.e., frequent data samples) and low-density (i.e., less frequent data samples) regions:

\begin{equation}
    \mathcal{L}_{adv} = \frac{1}{2} \left[ D_{KL}(P_{high} || M) + D_{KL}(P_{low} || M) \right],
\end{equation}
where $P_{high}$ and $P_{low}$ are the normalized histograms of median predictions for high and low-density regions (split by median value), $M = \frac{1}{2}(P_{high} + P_{low})$ is their average, and $D_{KL}$ is the Kullback-Leibler divergence:

\begin{equation}
    D_{KL}(P || Q) = \sum_{i} P_i \log\left(\frac{P_i}{Q_i}\right).
\end{equation}

This term encourages the model to produce similar prediction distributions regardless of whether the target values come from high or low-density regions, thereby reducing systematic bias due to the imbalanced training data.

\textbf{Total Loss:}

Total loss is the weighted sum of the above components:

\begin{equation}
    \begin{aligned} 
        \mathcal{L}_{\text{total}} &=
        \mathcal{L}_{\text{quantile}} +
        \lambda_{\text{mono}} \cdot \mathcal{L}_{\text{mono}} +
        \lambda_{\text{spatial}} \cdot \mathcal{L}_{\text{spatial}}\\ &\quad + \lambda_{\text{consistency}} \cdot \mathcal{L}_{\text{consistency}} + \lambda_{\text{adv}} \cdot \mathcal{L}_{\text{adv}}.
    \end{aligned}
\end{equation} 

The weights are adapted during training using a curriculum-based strategy. At initialization, the regularization weights are set to: $\lambda_{\text{mono}} = 0.01$, $\lambda_{\text{spatial}} = 0.001$, $\lambda_{\text{consistency}} = 0.001$, $\lambda_{\text{adv}} = 0.0$. The main prediction loss is assigned unit weight to ensure that early training is driven primarily by fitting the data, preventing premature over-regularisation when intermediate representations are still unstable. The weights are adapted in three phases during the training:

\textbf{1- Warm-up phase:}

During the initial stage, regularization terms are strongly downweighted. The model focuses on learning a stable predictive mapping from EO inputs, while constraints are only weakly enforced.

\textbf{2- Curriculum regularization phase:}

As training progresses, the weights of the regularization terms are gradually increased according to training progress. This introduces ecological and structural constraints smoothly, allowing intermediate concepts to adapt without disrupting convergence.

\textbf{3- Adaptive stabilization phase:}

In the final stage, regularization weights reach their maximum values of $\lambda_{\text{mono}} = 0.1$, $\lambda_{\text{spatial}} = 0.1$, $\lambda_{\text{consistency}} = 0.05$, $\lambda_{\text{adv}} = 0.01$. To avoid degradation in predictive performance, we monitor the validation RMSE and reduce all regularization weights if error increases. This adaptive mechanism balances constraint enforcement against predictive accuracy and prevents over-constraining the model.

Overall, this curriculum-based weighting strategy enables PG-CBM to first learn robust representations from sparse and heterogeneous data, and subsequently align them with ecological process knowledge without sacrificing accuracy or training stability. This adaptive weighting scheme can be interpreted as progressively shrinking the effective hypothesis space as training stabilizes, which allows process-guided constraints to act as a structural regulariser rather than a hard prior.

\section{Implementation and Evaluation Setup} \label{sup_section_Experimental Setup}

To ensure clarity and reproducibility, we describe here the experimental details used to implement and evaluate the proposed framework. This includes the training details and hyperparameters, description of the benchmark models (including vanilla CBM, equivalent black-box DL model and other existing AGBD products), the quantitative evaluation metrics, and other implementation details.

\subsection{Training details} \label{sup_section_Training details}

In the pre-training stage, each sub-model is trained for 500 epochs, and the model corresponding to the epoch with the lowest validation loss is selected. In the post-training stage, the complete PG-CBM framework is end-to-end fine-tuned for another 500 epochs, with the best-performing model similarly chosen based on validation loss. Processing patch size is 16$\times$16 pixels (25\,m pixel size) and all input and label data are normalised using their respective mean and standard deviation. We use a batch size of 32 and a learning rate of 1e-6, scheduled with Cosine Annealing and warm restarts \cite{loshchilov2016sgdr}. Optimisation is performed using the Adam optimiser, and gradient clipping is applied to prevent exploding gradients and ensure stable training.

\subsection{Benchmark 1: Vanilla CBM} \label{sup_section_vanilla CBM}

Standard CBMs as formulated in \cite{pmlr-v119-koh20a} cannot be directly applied to the AGBD estimation problem because they require complete $(x_i,z_i,y_i)$ tuples for training, where both concept and target labels are available for every sample. In our case, some concept labels (e.g., canopy cover and height) are available from GEDI data for $\sim$8.7 million patches, while some other concept labels (e.g., stem number density) and target labels (i.e., AGBD) are available from field-plots only for 8,098 patches. Since GEDI footprints and field plots rarely intersect, there are virtually no samples with both concept and AGBD labels. Consequently, a vanilla CBM cannot be trained as originally formulated. 

To approximate how a vanilla CBM would perform under our setting, we constrained the PG-CBM to behave as closely as possible to the classical CBM assumptions: 

\begin{itemize}
    \item \textbf{No process guidance:} The relationships among intermediate concepts (canopy cover, canopy height, stem number density) were removed, treating them as independent semantic attributes rather than causally linked process variables.
    
    \item \textbf{Jointly supervised subset:} To mimic the dense supervision requirement of vanilla CBM, we randomly sampled 8,098 GEDI-labelled patches (equal to the number of training field plot patches) and trained the model as if both concept and target labels were available for these samples. This enforces a pseudo joint supervision regime similar to a standard CBM’s complete $(x_i,z_i,y_i)$ dataset.
    
    \item \textbf{Single-source supervision:} All labels were treated as if originating from one homogeneous source, disregarding the true heterogeneous nature of GEDI and plot-derived annotations.
    
    \item \textbf{Static concept predictors:} After training the concept sub-models on the selected subset, their outputs were fixed and used as static inputs to the final AGBD prediction model. No joint fine-tuning or feedback between $h_i$ and $g$ was allowed, consistent with the two-stage training scheme of vanilla CBMs.
    
    \item \textbf{Concept independence:} The model architecture was restricted so that $g(\cdot)$ receives concept predictions as separate, uncorrelated inputs, preventing it from learning inter-concept causal relationships.
\end{itemize}

These constraints simulate how a vanilla CBM would operate given complete concept and target annotations and highlight its limitations under sparse, heterogeneous scientific supervision. They also illustrate why PG-CBM’s design (i.e., allowing partial, multi-source, and process-guided supervision) is essential for real-world applications such as AGBD estimation.

\subsection{Benchmark 2: Equivalent "black-box" DL model} \label{sup_section_Equivalent black-box DL model}

We compare the AGBD estimates from PG-CBM with an equivalent black-box DL model. The black-box DL model shares the same architecture as PG-CBM without the ecologically meaningful intermediate concepts, hence without the domain knowledge and interpretability. This model is trained using only the plot AGBD data as the concept labels (i.e., canopy cover and canopy height from GEDI and stem number density from plot data) are not accounted for in this model.

\subsection{Benchmark 3 and 4: State of the art AGBD products} \label{sup_section_Other AGBD products}

In order to compare the AGBD estimates with existing well-known AGBD maps, we extracted estimated AGBD for the validation plots from the European Space Agency (ESA) Climate Change Initiative (CCI) biomass map v5.01 \cite{santoro2024biomasscci} and the GEDI L4B product \cite{dubayah2022gedi}. Both CCI and GEDI biomass datasets were accessed through GEE. For ESA CCI, the AGBD values are taken from the map closest in time to each plot's census date, while for GEDI L4B, the mean AGBD of the 1 km$^2$ grid is used. Due to missing data caused by land cover masking in ESA CCI and the discontinuous nature of GEDI coverage, AGBD estimates are available from these sources for 123 and 110 validation plots, respectively. It should be noted that CCI and GEDI are global products, while the spatial domain of this study is focused on dry tropical woodlands of southern and central Africa. 

\subsection{Evaluation metrics} \label{sup_section_quantitaive_metrics_definition}

In order to quantitatively evaluate the AGBD estimates, we employ four commonly used error metrics in the CEOS biomass protocol \cite{duncanson2021aboveground}: Root Mean Square Deviation (RMSD), mean bias, mean absolute bias, and relative mean bias. These metrics capture overall accuracy, systematic error, average absolute deviation, and relative error, respectively:

\begin{equation}
    \text{RMSD} = \sqrt{\frac{1}{N} \sum_{i=1}^{N} ( \hat{y}_i - y_i )^2 },
\end{equation}

\begin{equation}
    \text{Mean Bias} = \frac{1}{N} \sum_{i=1}^{N} ( \hat{y}_i - y_i ),
\end{equation}

\begin{equation}
    \text{Mean Absolute Bias} = \frac{1}{N} \sum_{i=1}^{N} \left| \hat{y}_i - y_i \right|,
\end{equation}

\begin{equation}
\begin{aligned}
\text{Relative Mean Bias} =& \frac{\frac{1}{N} \sum_{i=1}^{N} ( \hat{y}_i - y_i )}{\frac{1}{N} \sum_{i=1}^{N}y_i} \times 100 \\&= \frac{\text{Mean Prediction Bias}}{\text{Mean Observed target}} \times 100.
\end{aligned}
\end{equation}

\bibliographystyle{IEEEtran} 
\bibliography{refs} 

\vfill